\documentclass[10pt,twocolumn,letterpaper]{article}

\usepackage[pagenumbers]{cvpr} % To force page numbers, e.g. for an arXiv version

\usepackage{amsmath,amsfonts,bm}

\def\eqref#1{equation~\ref{#1}}
\def\1{\bm{1}}

\def\vmu{{\bm{\mu}}}
\def\vtheta{{\bm{\theta}}}

\def\vepsilon{{\bm{\epsilon}}}

\def\vh{{\bm{h}}}

\def\vx{{\bm{x}}}

\def\mI{{\bm{I}}}

\DeclareMathAlphabet{\mathsfit}{\encodingdefault}{\sfdefault}{m}{sl}
\SetMathAlphabet{\mathsfit}{bold}{\encodingdefault}{\sfdefault}{bx}{n}

\def\gN{{\mathcal{N}}}

\def\sR{{\mathbb{R}}}

\newcommand{\E}{\mathbb{E}}

\newcommand{\R}{\mathbb{R}}

\usepackage{graphicx}
\usepackage{amsmath}
\usepackage{amssymb}
\usepackage{booktabs}
\usepackage{float}
\usepackage{makecell}
\usepackage{graphicx}
\usepackage{multirow}
\usepackage{colortbl}
\usepackage{xcolor}

\usepackage[pagebackref,breaklinks,colorlinks]{hyperref}

\usepackage[capitalize]{cleveref}
\crefname{section}{Sec.}{Secs.}
\Crefname{section}{Section}{Sections}
\Crefname{table}{Table}{Tables}
\crefname{table}{Tab.}{Tabs.}

\def\cvprPaperID{9022} % *** Enter the CVPR Paper ID here
\def\confName{CVPR}
\def\confYear{2023}

\begin{document}

\title{All are Worth Words: A ViT Backbone for Diffusion Models}

\author{
Fan Bao$^1$, Shen Nie$^2$, Kaiwen Xue$^2$, Yue Cao$^3$, Chongxuan Li$^2$\thanks{Corresponding to C. Li and J. Zhu.}~, Hang Su$^1$, Jun Zhu$^{1*}$ \\
$^1$Dept. of Comp. Sci. \& Tech., Institute for AI, BNRist Center \\
$^1$Tsinghua-Bosch Joint ML Center, THBI Lab,Tsinghua University, Beijing, 100084 China \\
$^2$Gaoling School of Artificial Intelligence, Renmin University of China, \\
$^2$Beijing Key Laboratory of Big Data Management and Analysis Methods, Beijing, China \\
$^3$Beijing Academy of Artificial Intelligence \\
{\tt\small bf19@mails.tsinghua.edu.cn;\ nieshen@ruc.edu.cn;\ \{kevin.kaiwenxue,\ caoyue10\}@gmail.com} \\
{\tt\small chongxuanli@ruc.edu.cn;\ \{suhangss,\ dcszj\}@tsinghua.edu.cn}
}
\maketitle

%%%%%%%%% ABSTRACT
\begin{abstract}
Vision transformers (ViT) have shown promise in various vision tasks while the U-Net based on a convolutional neural network (CNN) remains dominant in diffusion models. We design a simple and general ViT-based architecture (named U-ViT) for image generation with diffusion models. U-ViT is characterized by treating all inputs including the time, condition and noisy image patches as tokens and employing long skip connections between shallow and deep layers. We evaluate U-ViT in unconditional and class-conditional image generation, as well as text-to-image generation tasks, where U-ViT is comparable if not superior to a CNN-based U-Net of a similar size. In particular, latent diffusion models with U-ViT achieve record-breaking FID scores of 2.29 in class-conditional image generation on ImageNet 256$\times$256, and 5.48 in text-to-image generation on MS-COCO, among methods without accessing large external datasets during the training of generative models.

Our results suggest that, for diffusion-based image modeling, the long skip connection is crucial while the down-sampling and up-sampling operators in CNN-based U-Net are not always necessary. We believe that U-ViT can provide insights for future research on backbones in diffusion models and benefit generative modeling on large scale cross-modality datasets. Our code is available at \url{https://github.com/baofff/U-ViT}.
\end{abstract}

%%%%%%%%% BODY TEXT
\section{Introduction}

Diffusion models~\cite{sohl2015deep,ho2020denoising,song2020score} are powerful deep generative models that emerge recently for high quality image generation~\cite{dhariwal2021diffusion,rombach2022high,ho2022cascaded}. They grow rapidly and find applications in text-to-image generation~\cite{ramesh2022hierarchical,saharia2022photorealistic,rombach2022high}, image-to-image generation~\cite{zhao2022egsde,meng2021sdedit,choi2021ilvr}, video generation~\cite{ho2022video,ho2022imagen}, speech synthesis~\cite{chen2020wavegrad,kong2020diffwave}, and 3D synthesis~\cite{poole2022dreamfusion}.

\begin{figure}[t]
    \centering
    \includegraphics[width=\linewidth]{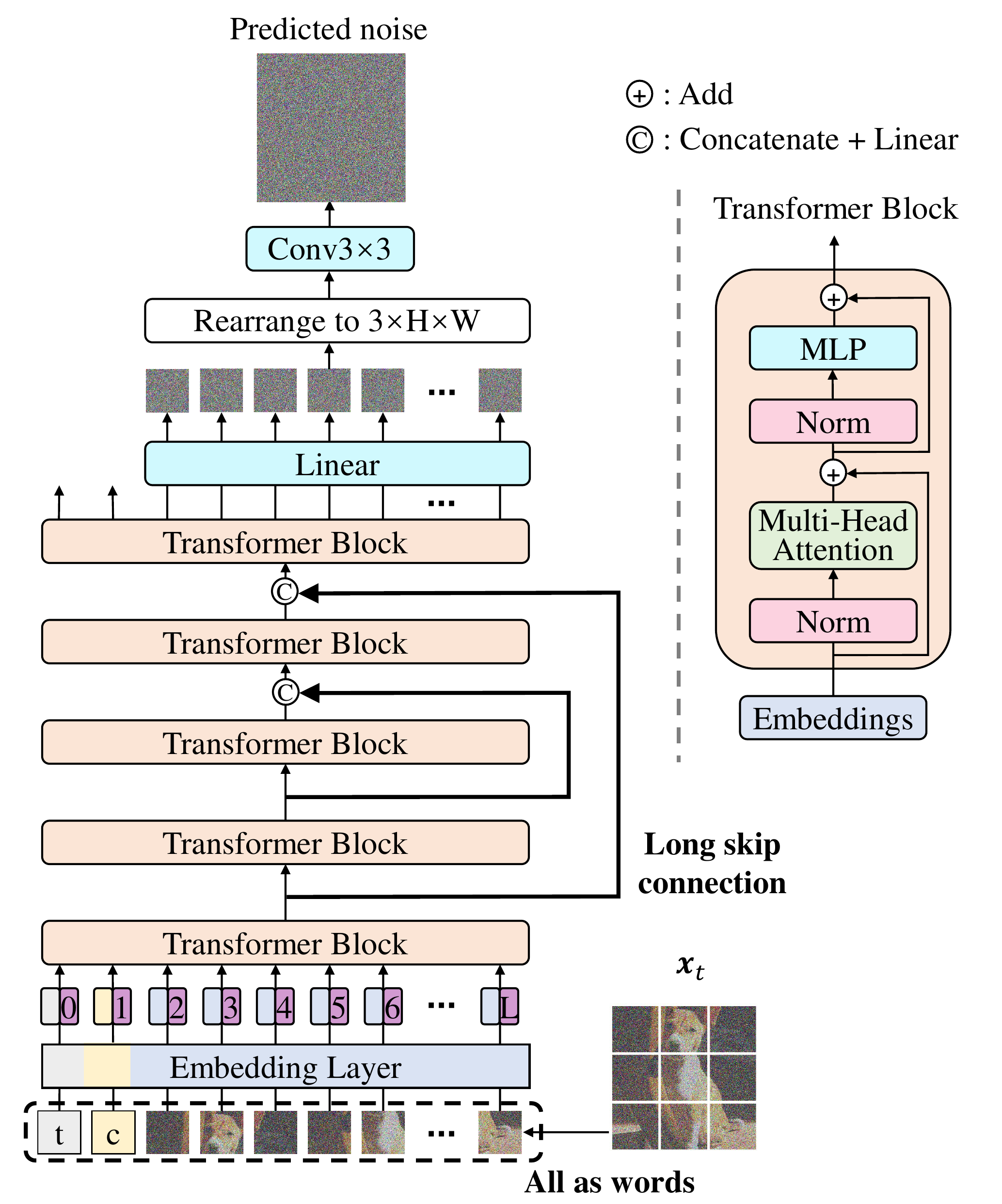}
    \caption{The \textbf{U-ViT} architecture for diffusion models, which is characterized by treating \textbf{all} inputs including the time, condition and noisy image patches \textbf{as tokens} and employing (\#Blocks-1)/2 \textbf{long skip connections} between shallow and deep layers.}\vspace{-.2cm}
    \label{fig:uvit}
\end{figure}

Along with the development of algorithms~\cite{ho2020denoising,song2020score,song2020denoising,nichol2021improved,song2021maximum,kingma2021variational,vahdat2021score,bao2022analytic,dockhorn2021score,bao2022estimating,lu2022maximum,lu2022dpm}, the revolution of backbones plays a central role in diffusion models. A representative example is U-Net based on a convolutional neural network (CNN) employed in prior work~\cite{song2019generative,ho2020denoising}. The CNN-based U-Net is characterized by a group of down-sampling blocks, a group of up-sampling blocks, and long skip connections between the two groups, which dominates diffusion models for image generation tasks~\cite{dhariwal2021diffusion,rombach2022high,ramesh2022hierarchical,saharia2022photorealistic}.  
On the other hand, vision transformers (ViT)~\cite{dosovitskiy2020image} have shown promise in various vision tasks, where ViT is comparable or even superior to CNN based approaches~\cite{chen2021empirical,he2022masked,strudel2021segmenter,zheng2021rethinking,lee2021vitgan}.
Therefore, a very natural question arises: \emph{whether the reliance of the CNN-based U-Net is necessary in diffusion models?}

\begin{figure*}[t]
\centering
\subfloat[Combine the long skip branch]{\includegraphics[width=0.3\linewidth]{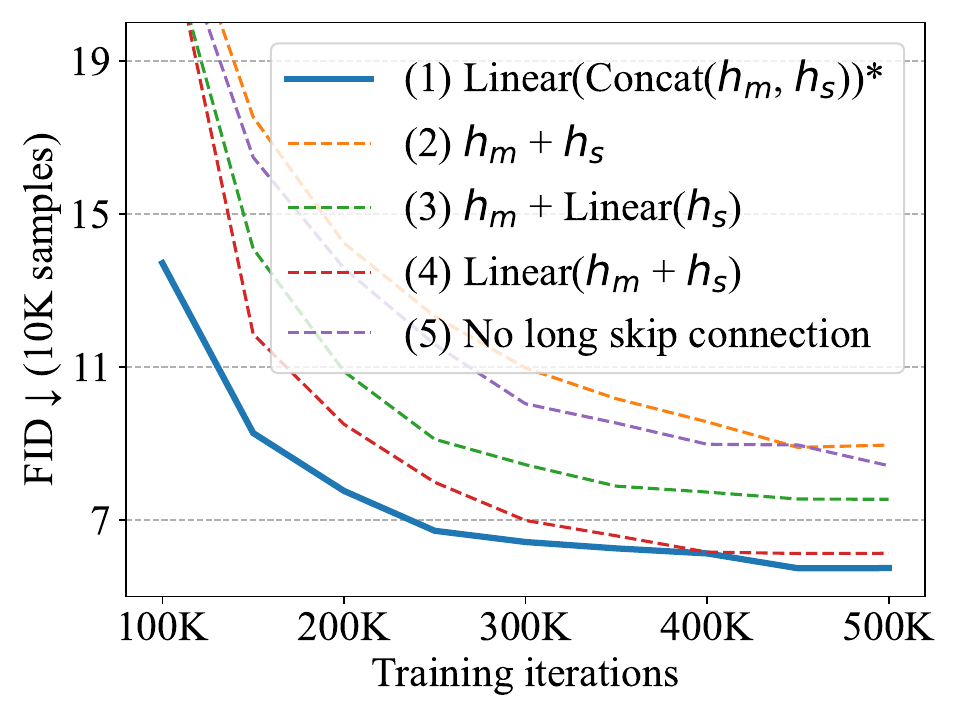}}
\hspace{.1cm}
\subfloat[Feed time into the network]{\includegraphics[width=0.3\linewidth]{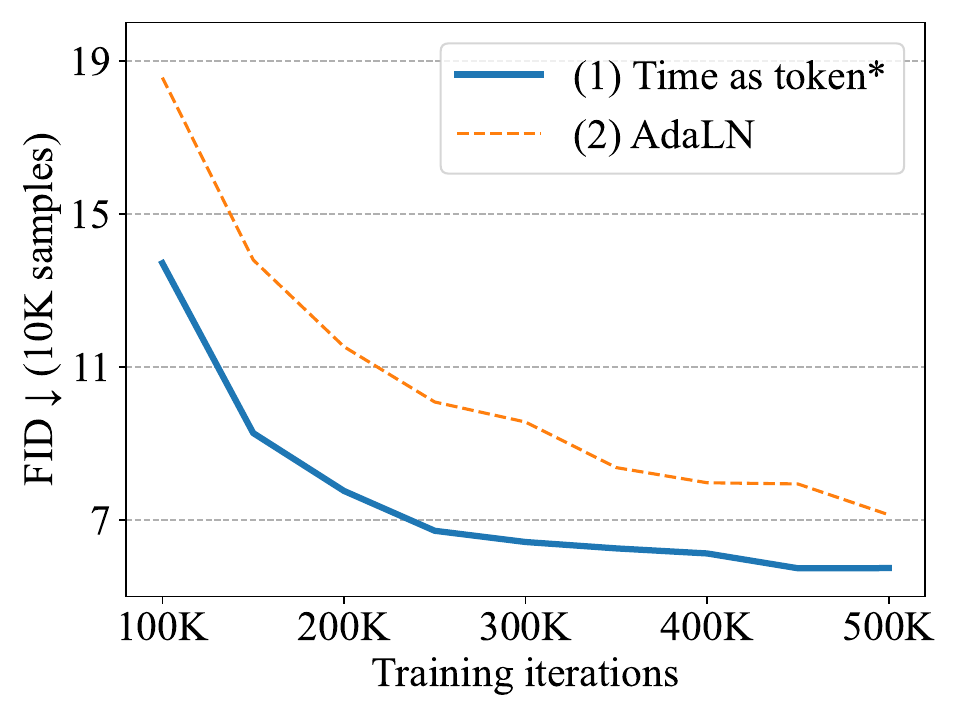}}
\hspace{.1cm}
\subfloat[Add an extra convolutional block]{\includegraphics[width=0.3\linewidth]{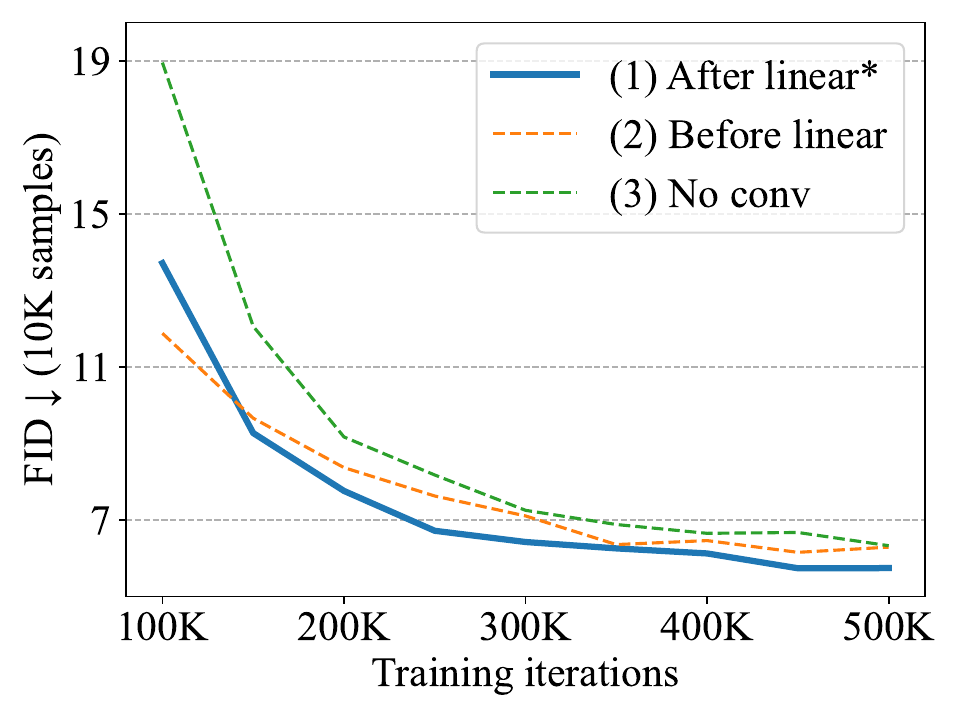}}\\
\vspace{.2cm}
\subfloat[Variants of patch embedding]{\includegraphics[width=0.3\linewidth]{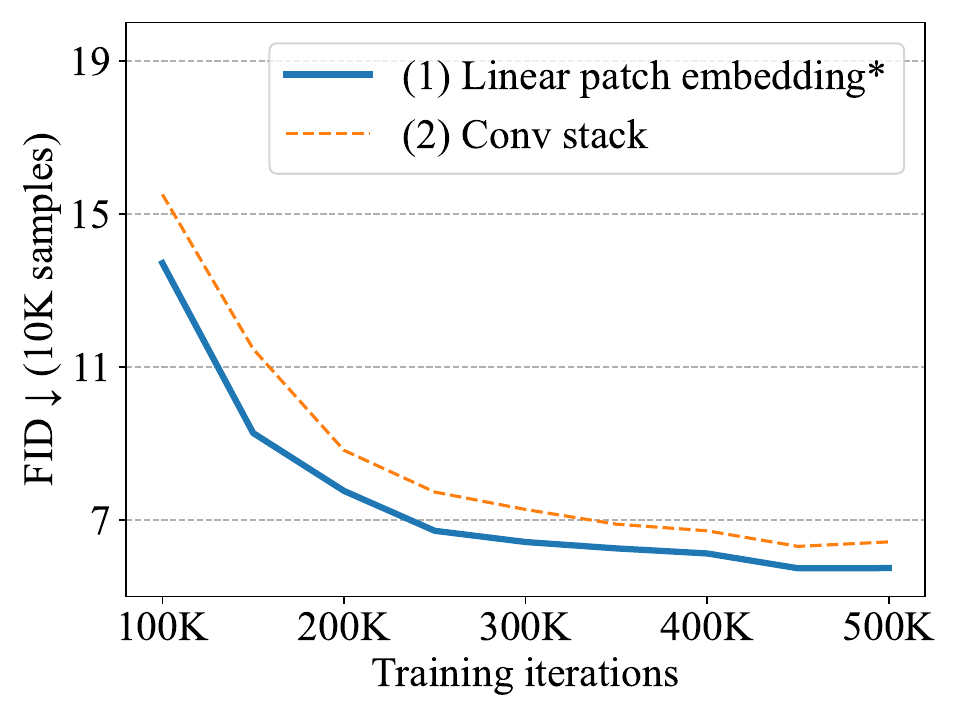}}
\hspace{.1cm}
\subfloat[Variants of position embedding]{\includegraphics[width=0.3\linewidth]{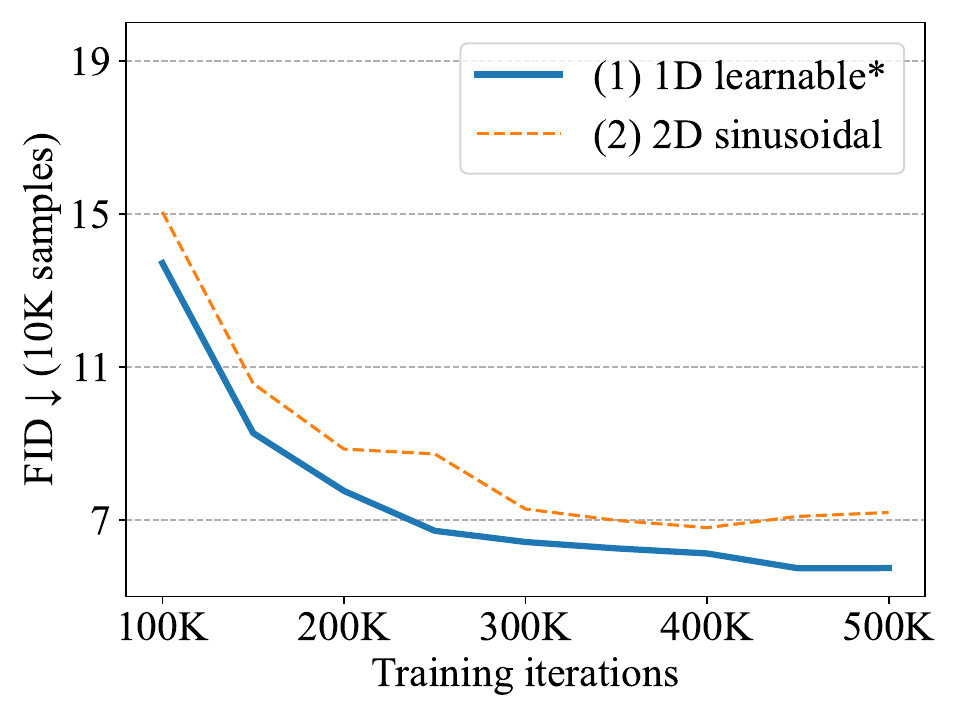}}
\caption{Ablate design choices. The one marked with * is the adopted choice of U-ViT illustrated in Figure~\ref{fig:uvit}. Since this ablation aims to determine implementation details, we evaluate FID on 10K generated samples (instead of 50K samples for efficiency).}
\label{fig:ab}
\end{figure*}

In this paper, we design a simple and general ViT-based architecture called U-ViT (Figure~\ref{fig:uvit}). Following the design methodology of transformers, U-ViT treats all inputs including the time, condition and noisy image patches as tokens. Crucially, U-ViT employs long skip connections between shallow and deep layers inspired by U-Net. Intuitively, low-level features are important to the pixel-level prediction objective in diffusion models and such connections can ease the training of the corresponding prediction network. Besides, U-ViT optionally adds an extra 3$\times$3 convolutional block before output for better visual quality.
See a systematical ablation study for all elements in Figure~\ref{fig:ab}.

We evaluate U-ViT in three popular tasks: unconditional image generation, class-conditional image generation and text-to-image generation. In all settings, U-ViT is comparable if not superior to a CNN-based U-Net of a similar size. 
In particular, latent diffusion models with U-ViT achieve record-breaking FID scores of 2.29 in class-conditional image generation on ImageNet 256$\times$256, and 5.48 in text-to-image generation on MS-COCO, among methods without accessing large external datasets during the training of generative models.

Our results suggest that the long skip connection is crucial while the down/up-sampling operators in CNN-based U-Net are not always necessary for image diffusion models. We believe that U-ViT can provide insights for future research on diffusion model backbones and benefit generative modeling on large scale cross-modality datasets.

\section{Background}

\textbf{Diffusion models}~\cite{sohl2015deep,ho2020denoising,song2020score} gradually inject noise to data, and then reverse this process to generate data from noise. The noise-injection process, also called the forward process, is formalized as a Markov chain:
\begin{align*}
    q(\vx_{1:T}|\vx_0) = \prod_{t=1}^T q(\vx_t|\vx_{t-1}).
\end{align*}
Here $\vx_0$ is the data, $q(\vx_t|\vx_{t-1}) = \gN(\vx_t|\sqrt{\alpha_t} \vx_{t-1}, \beta_t \mI)$, and $\alpha_t$ and $\beta_t$ represent the noise schedule such that $\alpha_t+\beta_t=1$. To reverse this process, a Gaussian model $p(\vx_{t-1}|\vx_t) = \gN(\vx_{t-1}|\vmu_t(\vx_t), \sigma_t^2 \mI)$ is adopted to approximate the ground truth reverse transition $q(\vx_{t-1}|\vx_t)$, and the optimal mean~\cite{bao2022analytic} is
\begin{align*}
    \vmu_t^*(\vx_t) = \frac{1}{\sqrt{\alpha_t}} \left(\vx_t - \frac{\beta_t}{\sqrt{1-\overline{\alpha}_t}} \E[\vepsilon|\vx_t]\right).
\end{align*}
Here $\overline{\alpha}_t = \prod_{i=1}^t \alpha_i$, and $\vepsilon$ is the standard Gaussian noises injected to $\vx_t$. Thus, the learning is equivalent to a noise prediction task. Formally, a noise prediction network $\vepsilon_\vtheta(\vx_t, t)$ is adopted to learn $\E[\vepsilon|\vx_t]$ by minimizing a noise prediction objective, i.e., $\min\limits_\vtheta \E_{t, \vx_0, \vepsilon} \| \vepsilon - \vepsilon_\vtheta(\vx_t, t) \|_2^2$, where $t$ is uniform between $1$ and $T$. To learn conditional diffusion models, e.g., class-conditional~\cite{dhariwal2021diffusion} or text-to-image~\cite{ramesh2022hierarchical} models, the condition information is further fed into the noise prediction objective:
\begin{align}
\label{eq:obj}
    \min\limits_\vtheta \E_{t, \vx_0, c, \vepsilon} \| \vepsilon - \vepsilon_\vtheta(\vx_t, t, c) \|_2^2,
\end{align}
where $c$ is the condition or its continuous embedding. In prior work on image modeling, the success of diffusion models heavily rely on CNN-based U-Net~\cite{ronneberger2015u,song2019generative}, which is a convolutional backbone characterized by a group of down-sampling blocks, a group of up-sampling blocks and long skip connections between the two groups, and $c$ is fed into U-Net by mechanisms such as adaptive group normalization~\cite{dhariwal2021diffusion} and cross attention~\cite{rombach2022high}.

\textbf{Vision Transformer (ViT)}~\cite{dosovitskiy2020image} is a pure transformer architecture that treats an image as a sequence of tokens (words). ViT rearranges an image into a sequence of flattened patches. Then, ViT adds learnable 1D position embeddings to linear embeddings of these patches before feeding them into a transformer encoder~\cite{vaswani2017attention}. ViT has shown promise in various vision tasks but it is not clear whether it is suitable for diffusion-based image modeling yet.

\section{Method}

U-ViT is a simple and general backbone for diffusion models in image generation (Figure~\ref{fig:uvit}). In particular,
U-ViT parameterizes the noise prediction network\footnote{U-ViT can also parameterize other types of prediction, e.g., $\vx_0$-prediction~\cite{ho2020denoising}.} $\vepsilon_\vtheta(\vx_t, t, c)$ in Eq.~{(\ref{eq:obj})}. It takes the time $t$, the condition $c$ and the noisy image $\vx_t$ as inputs and predicts the noise injected into $\vx_t$.
Following the design methodology of ViT, the image is split into patches, and U-ViT treats all inputs including the time, condition and image patches as tokens (words).

Inspired by the success of the CNN-based U-Net in diffusion models~\cite{song2019generative}, U-ViT also employs similar long skip connections between shallow and deep layers. Intuitively, the objective in Eq.~{(\ref{eq:obj})} is a pixel-level prediction task and is sensitive to low-level features. The long skip connections provide shortcuts for the low-level features and therefore ease the training of the noise prediction network.

Additionally, U-ViT optionally adds a 3$\times$3 convolutional block before output. This is intended to prevent the potential artifacts in images produced by transformers~\cite{zhang2022styleswin}. The block improves the visual quality of the samples generated by U-ViT according to our experiments.

In Section~\ref{sec:imp}, we present the implementation details of U-ViT. In Section~\ref{sec:ab2}, we present the scaling properties of U-ViT by studying the effect of depth, width and patch size.

\subsection{Implementation Details}
\label{sec:imp}
Although U-ViT is conceptually simple, we carefully design its implementation. To this end, we perform a systematical empirical study on key elements in U-ViT. In particular, we ablate on CIFAR10~\cite{krizhevsky2009learning}, evaluate the FID score~\cite{heusel2017gans} every 50K training iterations on 10K generated samples (instead of 50K samples for efficiency), and determine default implementation details.

\begin{figure*}[t]
\centering
\subfloat[Depth (\#layers)]{\includegraphics[width=0.3\linewidth]{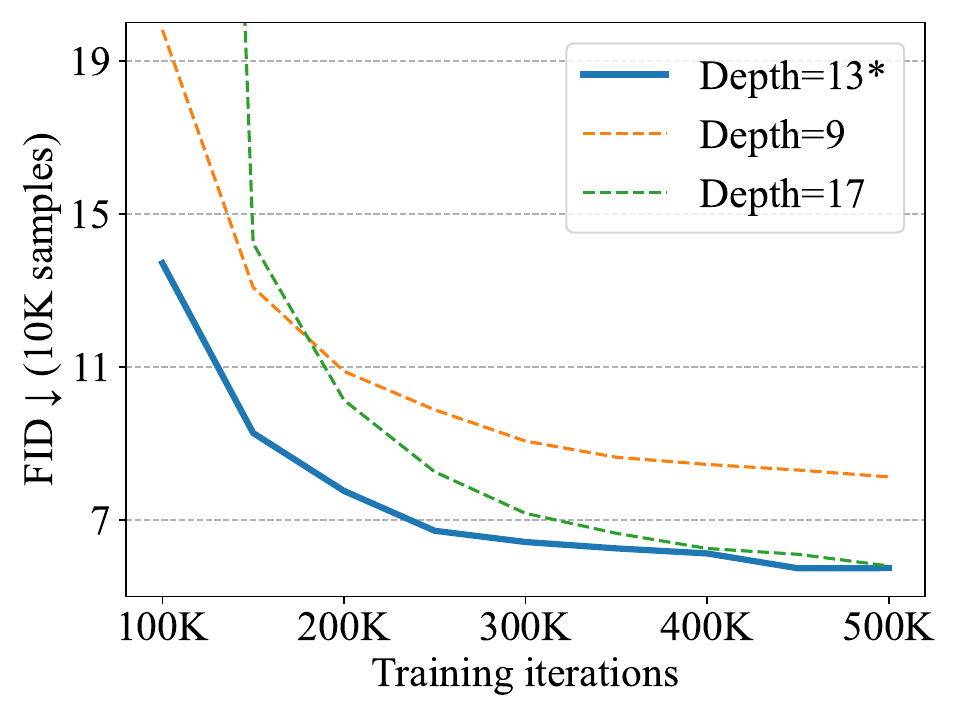}}
\hspace{.1cm}
\subfloat[Width (hidden size)]{\includegraphics[width=0.3\linewidth]{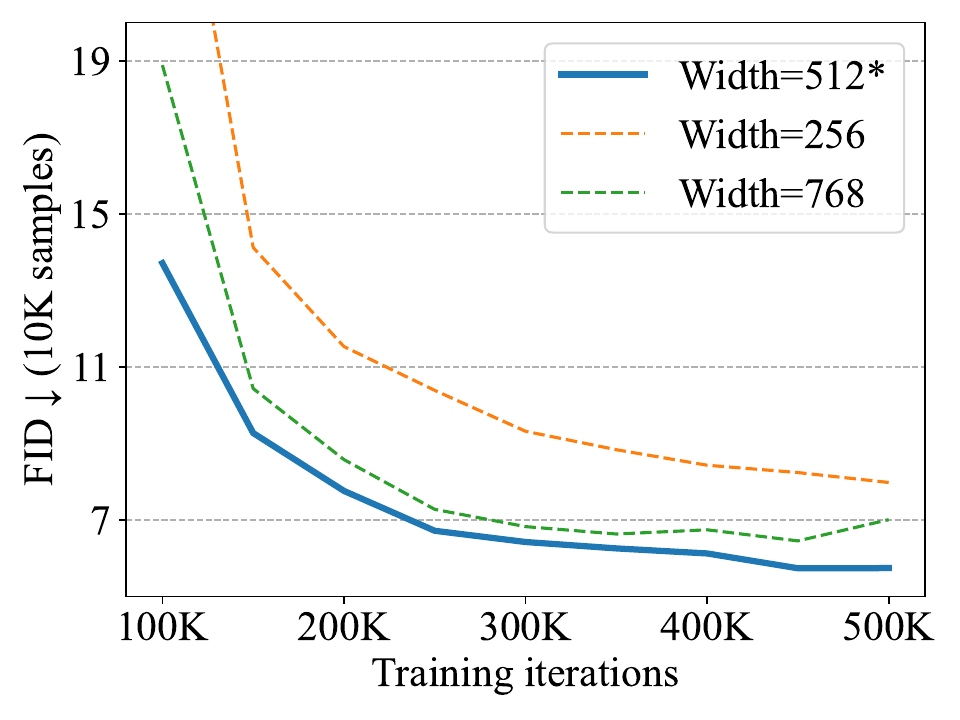}}
\hspace{.1cm}
\subfloat[Patch size]{\includegraphics[width=0.3\linewidth]{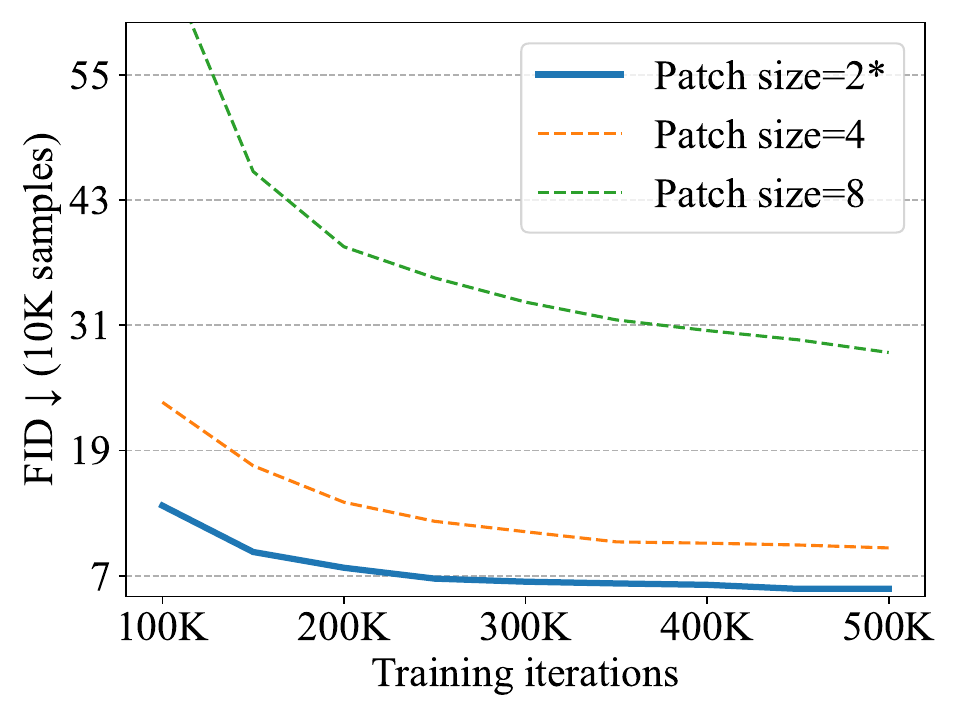}}
\vspace{-.2cm}
\caption{Effect of depth, width and patch size. The one marked with * corresponds to the setting of U-ViT-S/2 (see Table~\ref{tab:uvit_cfg}).}
\vspace{-.4cm}
\label{fig:ab2}
\end{figure*}

\textbf{The way to combine the long skip branch.} Let $\vh_m, \vh_s \in \sR^{L\times D}$ be the embeddings from the main branch and the long skip branch respectively. We consider several ways to combine them before feeding them to the next transformer block: (1) concatenating them and then performing a linear projection as illustrated in Figure~\ref{fig:uvit}, i.e., $\mathrm{Linear}(\mathrm{Concat}(\vh_m, \vh_s))$; (2) directly adding them, i.e., $\vh_m + \vh_s$; (3) performing a linear projection to $\vh_s$ and then adding them, i.e., $\vh_m + \mathrm{Linear}(\vh_s)$; (4) adding them and then performing a linear projection, i.e., $\mathrm{Linear}(\vh_m + \vh_s)$. (5) We also compare with the case where the long skip connection is dropped. As shown in Figure~{\ref{fig:ab} (a)}, directly adding $\vh_m, \vh_s$ does not provide benefits. Since a transformer block has skip connections via the adding operator inside it, $\vh_m$ already contains information of $\vh_s$ in a linear form. As a result, the only effect of $\vh_m + \vh_s$ is to increases the coefficient of $\vh_s$ in the linear form, which does not change the nature of the network. In contrast, all other ways to combine $\vh_s$ perform a linear projection on $\vh_s$ and improve the performance compared to no long skip connection. Among them, the first way with concatenation performs the best. In Appendix~\ref{app:cka}, we visualize the similarity between representations in a network, and we find the first way with concatenation significantly changes the representations, which validates its effectiveness.

\textbf{The way to feed the time into the network.} We consider two ways to feed $t$ into the network. (1) The first way is to treat it as a token as illustrated in Figure~\ref{fig:uvit}. (2) The second way is to incorporate the time after the layer normalization in the transformer block~\cite{gu2022vector}, which is similar to the adaptive group normalization~\cite{dhariwal2021diffusion} used in U-Net. The second way is referred to as adaptive layer normalization (AdaLN). Formally, $\mathrm{AdaLN}(h, y) = y_s \mathrm{LayerNorm}(h) + y_b$, where $h$ is an embedding inside a transformer block, and $y_s, y_b$ are obtained from a linear projection of the time embedding. As shown in Figure~{\ref{fig:ab} (b)}, while simple, the first way that treats time as a token performs better than AdaLN.

\textbf{The way to add an extra convolutional block after the transformer.} We consider two ways to add an extra convolutional block after the transformer. (1) The first way is to add a 3$\times$3 convolutional block after the linear projection that maps the token embeddings to image patches, as illustrated in Figure~\ref{fig:uvit}. (2) The second way is to add a 3$\times$3 convolutional block before this linear projection, which needs to first rearrange the 1D sequence of token embeddings $\vh \in \R^{L \times D}$ to a 2D feature of shape $H/P \times W/P \times D$, where $P$ is the patch size. (3) We also compare with the case where we drop the extra convolutional block. As shown in Figure~{\ref{fig:ab} (c)}, the first way that adds a 3$\times$3 convolutional block after the linear projection performs slightly better than other two choices.

\textbf{Variants of the patch embedding.} We consider two variants of the patch embedding. (1) The original patch embedding adopts a linear projection that maps a patch to a token embedding, as illustrated in Figure~\ref{fig:uvit}. (2) Alternatively, \cite{xiao2021early} use a stack of 3$\times$3 convolutional blocks followed by a 1$\times$1 convolutional block to map an image to token embeddings. We compare them in Figure~{\ref{fig:ab} (d)}, and the original patch embedding performs better.

\textbf{Variants of the position embedding.} We consider two variants of the position embedding. (1) The first one is the 1-dimensional learnable position embedding proposed in the original ViT~\cite{dosovitskiy2020image}, which is the default setting in this paper. (2) The second one is the 2-dimensional sinusoidal position embedding, which is obtained by concatenating the sinusoidal embeddings~\cite{vaswani2017attention} of $i$ and $j$ for a patch at position $(i,j)$. As shown in Figure~{\ref{fig:ab} (e)}, the 1-dimensional learnable position embedding performs better. We also try not use any position embedding, and find the model fails to generate meaningful images, which implies the position information is critical in image generation.

\subsection{Effect of Depth, Width and Patch Size}
\label{sec:ab2}

We present scaling properties of U-ViT by studying the effect of the depth (i.e., the number of layers), width (i.e., the hidden size $D$) and patch size on CIFAR10. As shown in Figure~\ref{fig:ab2}, the performance improves as the depth (i.e., the number of layers) increases from 9 to 13. Nevertheless, U-ViT does not gain from a larger depth like 17 in 50K training iterations. Similarly, increasing the width (i.e., the hidden size) from 256 to 512 improves the performance, and further increase to 768 brings no gain; decreasing the patch size from 8 to 2 improves the performance, and further decrease to 1 brings no gain. Note that a small patch size like 2 is required for a good performance. We hypothesize it is because that the noise prediction task in diffusion models is low-level and requires small patches, differing from high-level tasks (e.g., classification). Since using a small patch size is costly for high resolution images, we firstly convert them to low-dimensional latent representations~\cite{rombach2022high} and model these latent representations using U-ViT.

\begin{figure*}[t]
\centering
\subfloat[ImageNet 512$\times$512]{\includegraphics[width=0.64\linewidth]{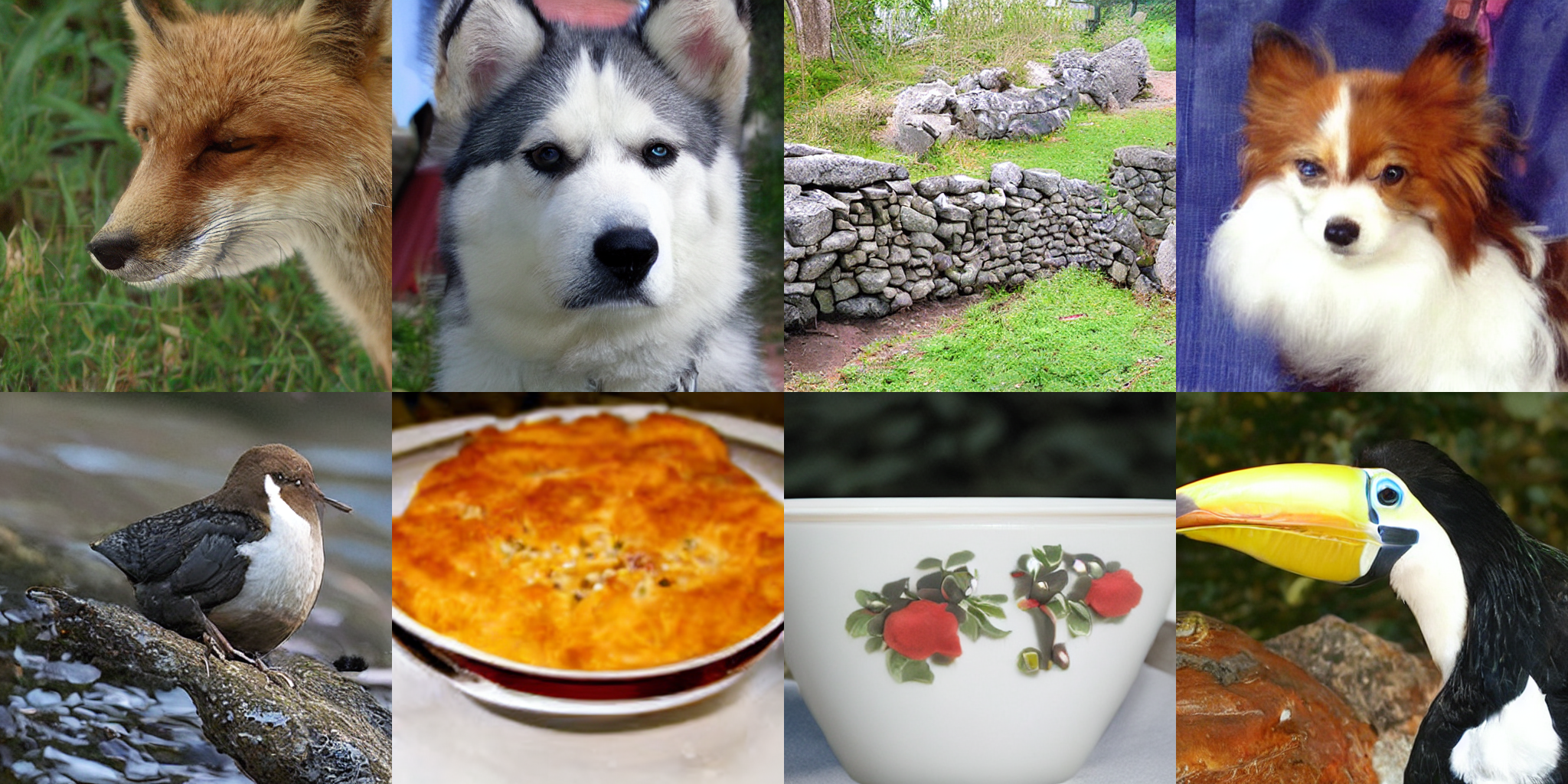}}
\hspace{.1cm}
\subfloat[ImageNet 256$\times$256]{\includegraphics[width=0.32\linewidth]{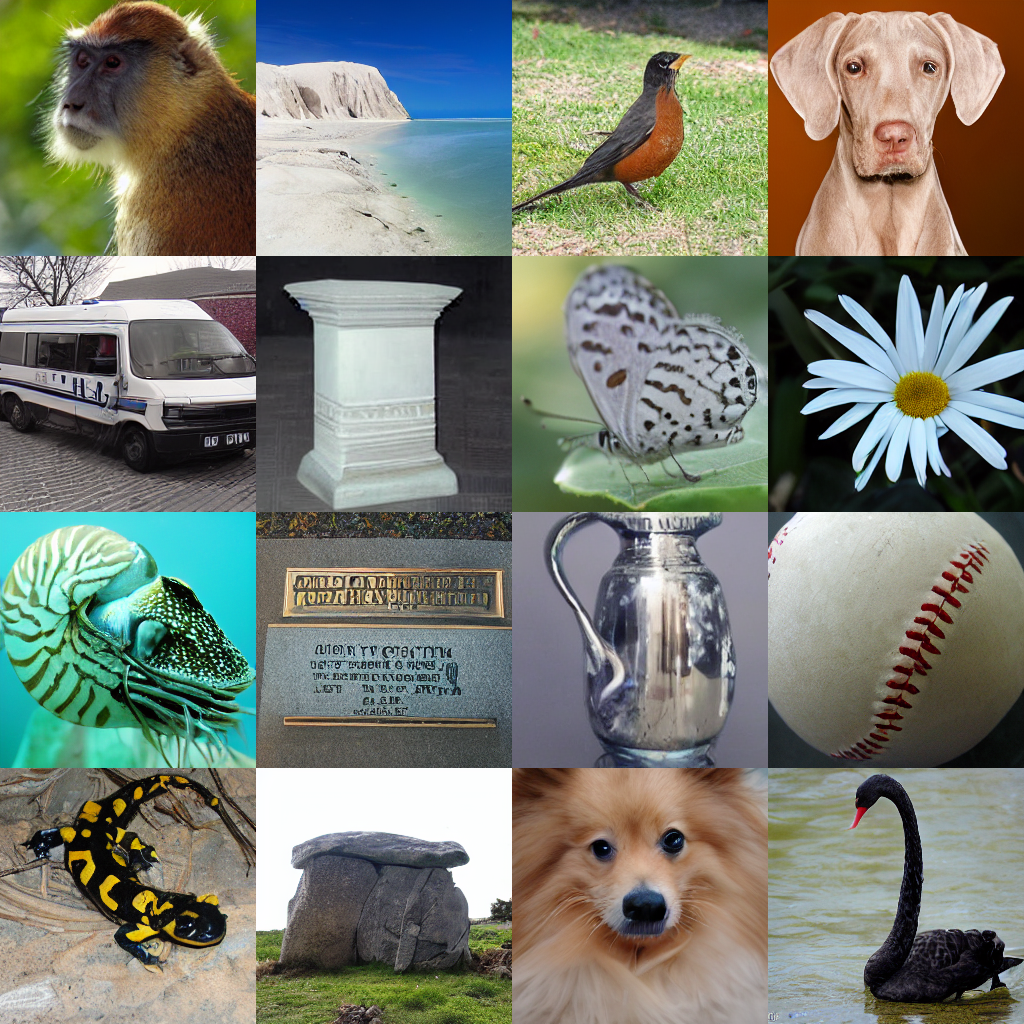}} \\
\vspace{.2cm}
\subfloat[CIFAR10]{\includegraphics[width=0.32\linewidth]{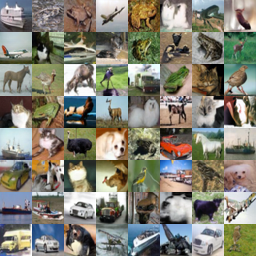}}
\hspace{.05cm}
\subfloat[CelebA 64$\times$64]{\includegraphics[width=0.32\linewidth]{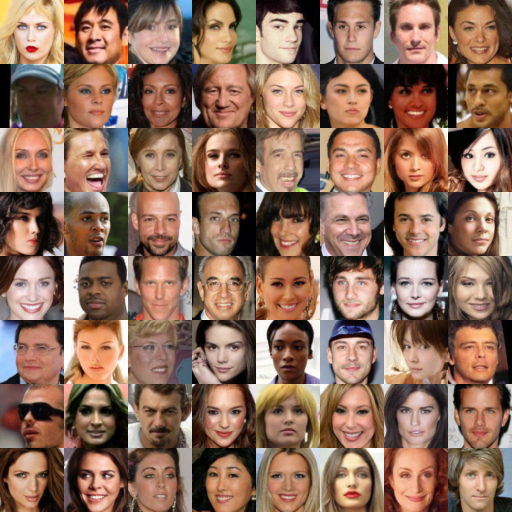}}
\hspace{.05cm}
\subfloat[ImageNet 64$\times$64]{\includegraphics[width=0.32\linewidth]{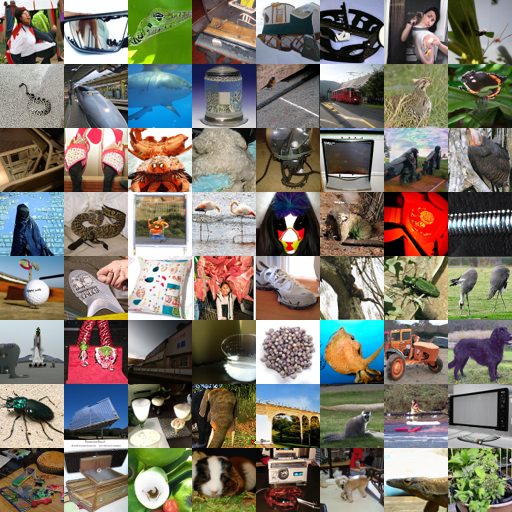}}\vspace{-.2cm}
\caption{Image generation results of U-ViT: selected samples on ImageNet 512$\times$512 and ImageNet 256$\times$256, and random samples on CIFAR10, CelebA 64$\times$64, and ImageNet 64$\times$64.}\vspace{-.2cm}
\label{fig:samples}
\end{figure*}

\section{Related Work}

\textbf{Transformers in diffusion models.} A related work is GenViT~\cite{yang2022your}. GenViT employs a smaller ViT that does not employ long skip connections and the 3$\times$3 convolutional block, and incorporates time before normalization layers for image diffusion models. Empirically, our U-ViT performs much better than GenViT (see Table~\ref{tab:fid}) by carefully designing implementation details. Another related work is VQ-Diffusion~\cite{gu2022vector} and its variants~\cite{tang2022improved,sheynin2022knn}. VQ-Diffusion firstly obtains a sequence of discrete image tokens via a VQ-GAN~\cite{esser2021taming}, and then models these tokens using a discrete diffusion model~\cite{sohl2015deep,austin2021structured} with a transformer as its backbone. Time and condition are fed into the transformer through cross attention or adaptive layer normalization. In contrast, our U-ViT simply treats all inputs as tokens, and employs long skip connections between shallow and deep layers, which achieves a better FID (see Table~\ref{tab:fid} and Table~\ref{tab:fid_ms}). In addition to images, transformers in diffusion models are also employed to encode texts~\cite{nichol2021glide,ramesh2022hierarchical,rombach2022high,saharia2022photorealistic}, decode texts~\cite{hoogeboom2021argmax,nachmani2021zero,chen2022analog,li2022diffusion} and generate CLIP embeddings~\cite{ramesh2022hierarchical}.

\textbf{U-Net in diffusion models.} \cite{song2019generative,song2020improved} initially introduce CNN-based U-Net to model the gradient of log-likelihood function for continuous image data. After that, improvements on the CNN-based U-Net for (continuous) image diffusion models are made, including using group normalization~\cite{ho2020denoising}, multi-head attention~\cite{nichol2021improved}, improved residual block~\cite{dhariwal2021diffusion} and cross attention~\cite{rombach2022high}. In contrast, our U-ViT is a ViT-based backbone with conceptually simple design, and meanwhile has a comparable performance if not superior to a CNN-based U-Net of a similar size (see Table~\ref{tab:fid} and Table~\ref{tab:fid_ms}).

\textbf{Improvements of diffusion models.} In addition to the backbone, there are also improvements on other aspects, such as fast sampling~\cite{song2020denoising,bao2022analytic,lu2022dpm,salimans2022progressive,xiao2021tackling}, improved training methodology~\cite{nichol2021improved,song2021maximum,bao2022estimating,lu2022maximum,dockhorn2021score,vahdat2021score,kim2022soft,karras2022elucidating,kingma2021variational} and controllable generation~\cite{dhariwal2021diffusion,ho2022classifier,zhao2022egsde,bao2022equivariant,meng2021sdedit,choi2021ilvr,sehwag2022generating,hertz2022prompt}.

\begin{table*}[t]
\vspace{.1cm}
\centering
\scalebox{0.8}{
\begin{tabular}{lcc}
\toprule
    \multicolumn{2}{l}{Model on \textbf{CIFAR10}} & FID $\downarrow$\\
\midrule
\multicolumn{3}{l}{\textcolor{gray}{GAN}} \\
\arrayrulecolor{black!30}\midrule
\multicolumn{2}{l}{\quad \textcolor{gray}{StyleGAN2-ADA}~\cite{karras2020training}} & \textcolor{gray}{2.92} \\
\arrayrulecolor{black}\midrule
Diff. based on U-Net & \#Params & \\
\arrayrulecolor{black!30}\midrule
    \quad DDPM~\cite{ho2020denoising} & 36M & 3.17 \\
    \quad IDDPM~\cite{nichol2021improved} & 53M & 2.90 \\
    \quad DDPM++ cont.~\cite{song2020score} & 62M & 2.55 \\
    \quad EDM$^\dagger$~\cite{karras2022elucidating} & 56M & \textbf{1.97} \\
\arrayrulecolor{black}\midrule
Diff. based on ViT & \#Params & \\
\arrayrulecolor{black!30}\midrule
    \quad GenViT~\cite{yang2022your} & 11M & 20.20 \\
    \quad U-ViT-S/2 & 44M & 3.11 \\
\arrayrulecolor{black}\bottomrule
\\ \\
\toprule
    \multicolumn{2}{l}{Model on \textbf{CelebA 64$\times$64}} & FID $\downarrow$ \\
\midrule
Diff. based on U-Net & \#Params \\
\arrayrulecolor{black!30}\midrule
    \quad DDIM~\cite{song2020denoising} & 79M & 3.26 \\
    \quad Soft Truncation$^\dagger$~\cite{kim2022soft} & 62M & \textbf{1.90} \\
\arrayrulecolor{black}\midrule
Diff. model based on ViT & \#Params \\
\arrayrulecolor{black!30}\midrule
    \quad U-ViT-S/4 & 44M & 2.87 \\
\arrayrulecolor{black}\bottomrule
\\ \\
\toprule
    \multicolumn{2}{l}{Model on \textbf{ImageNet 64$\times$64}} & FID $\downarrow$ \\
\midrule
\multicolumn{3}{l}{\textcolor{gray}{GAN}} \\
\arrayrulecolor{black!30}\midrule
    \multicolumn{2}{l}{\quad \textcolor{gray}{BigGAN-deep}~\cite{brock2018large}} & \textcolor{gray}{4.06} \\
    \multicolumn{2}{l}{\quad \textcolor{gray}{StyleGAN-XL}~\cite{sauer2022stylegan}} & \textcolor{gray}{1.51} \\
\arrayrulecolor{black}\midrule
Diff. based on U-Net & \#Params \\
\arrayrulecolor{black!30}\midrule
    \quad IDDPM (small)~\cite{nichol2021improved} & 100M & 6.92 \\
    \quad IDDPM (large)~\cite{nichol2021improved} & 270M & 2.92 \\
    \quad CDM~\cite{ho2022cascaded} & Unknown & 1.48 \\
    \quad ADM~\cite{dhariwal2021diffusion} & 296M & 2.07 \\
    \quad EDM$^\dagger$~\cite{karras2022elucidating} & 296M & \textbf{1.36} \\
\arrayrulecolor{black}\midrule
Diff. based on ViT & \#Params \\
\arrayrulecolor{black!30}\midrule
    \quad U-ViT-M/4 & 131M & 5.85 \\
    \quad U-ViT-L/4 & 287M & 4.26 \\
\arrayrulecolor{black}\bottomrule
\end{tabular}}\quad\quad
\scalebox{0.8}{
\begin{tabular}{lcc}
\toprule
    \multicolumn{2}{l}{Model on \textbf{ImageNet 256$\times$256}} & FID $\downarrow$  \\
\midrule
\multicolumn{3}{l}{\textcolor{gray}{GAN}} \\
\arrayrulecolor{black!30}\midrule
    \multicolumn{2}{l}{\quad \textcolor{gray}{BigGAN-deep}~\cite{brock2018large}} & \textcolor{gray}{6.95} \\
    \multicolumn{2}{l}{\quad \textcolor{gray}{StyleGAN-XL}~\cite{sauer2022stylegan}} & \textcolor{gray}{2.30} \\
\arrayrulecolor{black}\midrule
\multicolumn{3}{l}{Discrete diff. based on transformer} \\
\arrayrulecolor{black!30}\midrule
    \multicolumn{2}{l}{\quad VQ-Diffusion~\cite{gu2022vector}} & 11.89 \\
    \multicolumn{2}{l}{\quad VQ-Diffusion (acc0.05)~\cite{gu2022vector}} & 5.32 \\
\arrayrulecolor{black}\midrule
Diff. based on U-Net & \#Params \\
\arrayrulecolor{black!30}\midrule
    \quad IDDPM~\cite{nichol2021improved} & 270M + 280M (SR) & 12.26 \\
    \quad CDM~\cite{ho2022cascaded} & Unknown & 4.88 \\
    \quad ADM~\cite{dhariwal2021diffusion} & 554M & 10.94 \\
    \quad ADM-U~\cite{dhariwal2021diffusion} & 296M + 312M (SR) & 7.49 \\
    \quad ADM-G~\cite{dhariwal2021diffusion} & 554M + 54M (Cls) & 4.59 \\
    \quad ADM-G, ADM-U~\cite{dhariwal2021diffusion} & 296M + 65M (Cls) + 312M (SR)  & 3.94 \\
    \quad LDM$^\ddagger$~\cite{rombach2022high} & 400M + 55M (AE) & 3.60 \\
\arrayrulecolor{black}\midrule
Diff. based on ViT & \#Params \\
\arrayrulecolor{black!30}\midrule
    \quad U-ViT-H/2$^\ddagger$ & 501M + 84M (AE) & \textbf{2.29} \\
\arrayrulecolor{black}\bottomrule
\\ \\
\toprule
    \multicolumn{2}{l}{Model on \textbf{ImageNet 512$\times$512}} & FID $\downarrow$ \\
\midrule
\multicolumn{3}{l}{\textcolor{gray}{GAN}} \\
\arrayrulecolor{black!30}\midrule
    \multicolumn{2}{l}{\quad \textcolor{gray}{BigGAN-deep}~\cite{brock2018large}} & \textcolor{gray}{8.43} \\
    \multicolumn{2}{l}{\quad \textcolor{gray}{StyleGAN-XL}~\cite{sauer2022stylegan}} & \textcolor{gray}{2.41} \\
\arrayrulecolor{black}\midrule
Diff. based on U-Net & \#Params \\
\arrayrulecolor{black!30}\midrule
    \quad ADM~\cite{dhariwal2021diffusion} & 559M & 23.24 \\
    \quad ADM-U~\cite{dhariwal2021diffusion} & 422M + 309M (SR) & 9.96 \\
    \quad ADM-G~\cite{dhariwal2021diffusion} & 559M + 54M (Cls) & 7.72 \\
    \quad ADM-G, ADM-U~\cite{dhariwal2021diffusion} & 422M + 43M (Cls) + 309M (SR) & \textbf{3.85} \\
\arrayrulecolor{black}\midrule
Diff. based on ViT & \#Params \\
\arrayrulecolor{black!30}\midrule
    \quad U-ViT-H/4$^\ddagger$ & 501M + 84M (AE) & 4.05 \\
\arrayrulecolor{black}\bottomrule
\end{tabular}}
\caption{FID results of unconditional image generation on CIFAR10 and CelebA 64$\times$64, and class-conditional image generation on ImageNet 64$\times$64, 256$\times$256 and 512$\times$512. We mark the best results \textit{among diffusion models} in \textbf{bold}. We also include GAN results (\textcolor{gray}{gray}) for completeness. Methods marked with $^\dagger$ use advanced training techniques for diffusion models. Methods marked with $^\ddagger$ model latent representations of images~\cite{rombach2022high} and use classifier-free guidance~\cite{ho2022classifier}. We also present the number of parameters of auxiliary components for diffusion models, where SR represents a super-resolution module, AE represents an image autoencoder, and Cls represents a classifier.}\vspace{-.2cm}
\label{tab:fid}
\end{table*}

\section{Experiments}

We evaluate the proposed U-ViT in unconditional and class-conditional image generation (Section~\ref{sec:gen}), as well as text-to-image generation (Section~\ref{sec:t2i}). Before presenting these results, we list main experimental setup below, and more details such as the sampling hyperparameters are provided in Appendix~\ref{app:setup}.

\subsection{Experimental Setup}
\label{sec:setup}

\textbf{Datasets.} For unconditional learning, we consider CIFAR10~\cite{krizhevsky2009learning}, which contain 50K training images, and CelebA 64$\times$64~\cite{liu2015faceattributes}, which contain 162,770 training images of human faces. For class-conditional learning, we consider ImageNet~\cite{deng2009imagenet} at 64$\times$64, 256$\times$256 and 512$\times$512 resolutions, which contains 1,281,167 training images from 1K different classes. For text-to-image learning, we consider MS-COCO~\cite{lin2014microsoft} at 256$\times$256 resolution, which contains 82,783 training images and 40,504 validation images. Each image is annotated with 5 captions.

\textbf{High resolution image generation.} We follow latent diffusion models (LDM)~\cite{rombach2022high} for images at 256$\times$256 and 512$\times$512 resolutions. We firstly convert them to latent representations at 32$\times$32 and 64$\times$64 resolutions respectively, using a pretrained image autoencoder provided by Stable Diffusion\footnote{\label{sd}\url{https://github.com/CompVis/stable-diffusion}}~\cite{rombach2022high}. Then we model these latent representations using the proposed U-ViT.

\textbf{Text-to-image learning.} On MS-COCO, we convert discrete texts to a sequence of embeddings using a CLIP text encoder following Stable Diffusion. Then these embeddings are fed into U-ViT as a sequence of tokens.

\textbf{U-ViT configurations.} We identify several configurations of U-ViT in Table~\ref{tab:uvit_cfg}. In the rest of the paper, we use brief notation to represent the U-ViT configuration and the input patch size (for instance, U-ViT-H/2 means the U-ViT-Huge configuration with an input patch size of 2$\times$2).
\begin{table}[H]
    \centering
    \vspace{-.2cm}
    \scalebox{0.67}{
    \begin{tabular}{lccccc}
    \toprule
        Model & \#Layers & Hidden size $D$ & MLP size & \#Heads & \#Params \\
    \midrule
        U-ViT-Small & 13 & 512 & 2048 & 8 & 44M \\
        U-ViT-Small (Deep) & 17 & 512 & 2048 & 8 & 58M \\
        U-ViT-Mid & 17 & 768 & 3072 & 12 & 131M \\
        U-ViT-Large & 21 & 1024 & 4096 & 16 & 287M \\
        U-ViT-Huge & 29 & 1152 & 4608 & 16 & 501M \\
    \bottomrule
    \end{tabular}}\vspace{-.2cm}
    \caption{Configurations of U-ViT.}\vspace{-.3cm}
    \label{tab:uvit_cfg}
\end{table}

\textbf{Training.} We use the AdamW optimizer~\cite{loshchilov2017decoupled} with a weight decay of 0.3 for all datasets. We use a learning rate of 2e-4 for most datasets, except ImageNet 64$\times$64 where we use 3e-4.
We train 500K iterations on CIFAR10 and CelebA 64$\times$64 with a batch size of 128. We train 300K iterations on ImageNet 64$\times$64 and ImageNet 256$\times$256, and 500K iterations on ImageNet 512$\times$512, with a batch size of 1024. We train 1M iterations on MS-COCO with a batch size of 256. On ImageNet 256$\times$256, ImageNet 512$\times$512 and MS-COCO, we adopt classifier-free guidance~\cite{ho2022classifier} following \cite{rombach2022high}.
We provide more details, such as the training time and how we choose hyperparameters in Appendix~\ref{app:setup}.

\begin{table}[t]
\centering
\scalebox{0.83}{\begin{tabular}{lrrrrrrr}
\toprule
 & 4 & 5 & 10 & 15 & 20  \\
\arrayrulecolor{black}\midrule
LDM (trained 178K) & 34.48 & 12.73 & 4.51 & 3.87 & 3.68  \\
\arrayrulecolor{black!30}\midrule
U-ViT-H/2 (trained 200K)  & 16.48 & 4.94 & 3.87 & 3.54  & 2.91  \\
\arrayrulecolor{black!30}\midrule
U-ViT-H/2 (trained 500K)  & 15.44 & 4.64 & 3.18 & 2.92 & 2.53  \\
\arrayrulecolor{black}\bottomrule
\end{tabular}}
\caption{FID results on ImageNet 256$\times$256 under different number of sampling steps using the DPM-Solver sampler~\cite{lu2022dpm}.}\vspace{-.2cm}
\label{tab:steps}
\end{table}

\begin{figure}[t]
    \centering
    \includegraphics[width=0.85\linewidth]{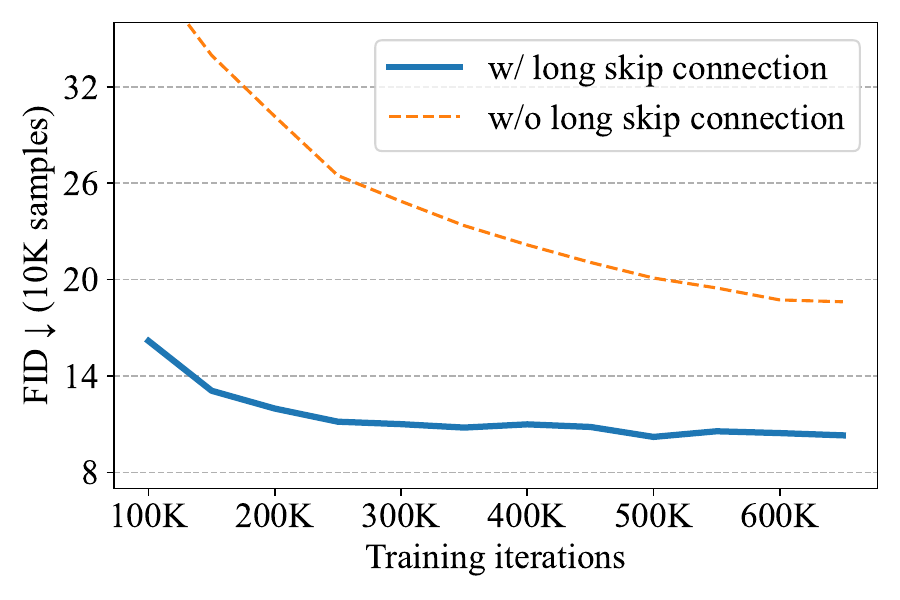}
    \caption{Ablate the long skip connection on ImageNet 256$\times$256 (w/o classifier-free guidance).}\vspace{-.2cm}
    \label{fig:skip_im}
\end{figure}

\begin{table*}[t]
    \centering
    \scalebox{0.85}{
    \begin{tabular}{lcccc}
    \toprule
        Model & FID & Type & Training datasets & \#Params \\
    \midrule
    \multicolumn{4}{l}{Generative model trained on external large dataset (zero-shot)} \\
    \arrayrulecolor{black!30}\midrule
        \quad DALL-E~\cite{ramesh2021zero} & $\sim$ 28 & Autoregressive & DALL-E dataset (250M) & 12B \\
        \quad CogView~\cite{ding2021cogview} & 27.1 & Autoregressive & Internal dataset (30M) & 4B\\
        \quad LAFITE~\cite{zhou2021lafite} & 26.94 & GAN & CC3M (3M) & 75M + 151M (TE)  \\
        \quad GLIDE~\cite{nichol2021glide} & 12.24 & Diffusion & DALL-E dataset (250M)&  3.5B + 1.5B (SR) \\
        \quad Make-A-Scene~\cite{gafni2022make} & 11.84 & Autoregressive & Union datasets (without MS-COCO) (35M)& 4B\\
        \quad DALL-E 2~\cite{ramesh2022hierarchical} & 10.39 & Diffusion & DALL-E dataset (250M)& 4.5B + 700M (SR)\\
        \quad Imagen~\cite{saharia2022photorealistic} & 7.27 & Diffusion & Internal dataset (460M) + LAION (400M) & 2B + 4.6B (TE) + 600M (SR) \\
        \quad Parti~\cite{yu2022scaling} & 7.23 & Autoregressive & LAION (400M) + FIT (400M) + JFT (4B) & 20B + 630M (AE) \\
        \quad Re-Imagen~\cite{chen2022re} & \textbf{6.88} & Diffusion & KNN-ImageText (50M) & 2.5B + 750M (SR) \\
    \arrayrulecolor{black}\midrule
    \multicolumn{4}{l}{Generative model trained on external large dataset with access to MS-COCO} \\
    \arrayrulecolor{black!30}\midrule
        \quad VQ-Diffusion$^\dagger$~\cite{gu2022vector} & 13.86 & Discrete diffusion & Conceptual Caption Subset (7M) &  370M \\
        \quad Make-A-Scene~\cite{gafni2022make} & 7.55 & Autoregressive & Union datasets (with MS-COCO) (35M) & 4B \\
        \quad Re-Imagen$^\ddagger$~\cite{chen2022re} & 5.25 & Diffusion & KNN-ImageText (50M) & 2.5B + 750M (SR) \\
        \quad Parti$^\dagger$~\cite{yu2022scaling} & \textbf{3.22} & Autoregressive & LAION (400M) + FIT (400M) + JFT (4B) & 20B + 630M (AE) \\
    \arrayrulecolor{black}\midrule
    \multicolumn{4}{l}{Generative model trained on MS-COCO} \\
    \arrayrulecolor{black!30}\midrule
        \quad AttnGAN~\cite{xu2018attngan} & 35.49 & GAN &  MS-COCO (83K) & 230M\\
        \quad DM-GAN~\cite{zhu2019dm} & 32.64 & GAN & MS-COCO (83K) & 46M\\
        \quad VQ-Diffusion~\cite{gu2022vector} & 19.75 & Discrete diffusion & MS-COCO (83K) & 370M \\
        \quad DF-GAN~\cite{tao2022df} & 19.32 & GAN & MS-COCO (83K) & 19M  \\
        \quad XMC-GAN~\cite{zhang2021cross} & 9.33 & GAN & MS-COCO (83K) & 166M \\
        \quad Friro~\cite{fan2022frido} & 8.97 & Diffusion & MS-COCO (83K) & 512M + 186M (TE) + 68M (AE) \\
        \quad LAFITE~\cite{zhou2021lafite} & 8.12 & GAN & MS-COCO (83K) & 75M + 151M (TE)\\
    \arrayrulecolor{black!30}\midrule
        \quad U-Net$^*$ & 7.32 & Latent diffusion & MS-COCO (83K) & 53M + 123M (TE) + 84M (AE) \\
        \quad U-ViT-S/2 & 5.95 & Latent diffusion & MS-COCO (83K) & 45M + 123M (TE) + 84M (AE) \\
        \quad U-ViT-S/2 (Deep) & \textbf{5.48} & Latent diffusion & MS-COCO (83K) & 58M + 123M (TE) + 84M (AE) \\
    \arrayrulecolor{black}\bottomrule
    \end{tabular}}
    \caption{FID results of different models on MS-COCO validation ($256 \times 256$). U-ViT-S (Deep) increases the number of layers from 13 to 17 compared to U-ViT-S. We also present the number of parameters of auxiliary components for a model when it is reported in the corresponding paper, where SR represents a super-resolution module, AE represents an image autoencoder, and TE represents a text encoder. Methods marked with $^\dagger$ finetune on MS-COCO. Methods marked with $^\ddagger$ use MS-COCO as a knowledge base for retrieval. The U-Net$^*$ is trained by ourselves to serve as a direct baseline of U-ViT, where we leave other parts unchanged except for the backbone.}\label{tab:fid_ms}
\end{table*}

\begin{figure*}[t]
    \centering
    \includegraphics[width=\linewidth]{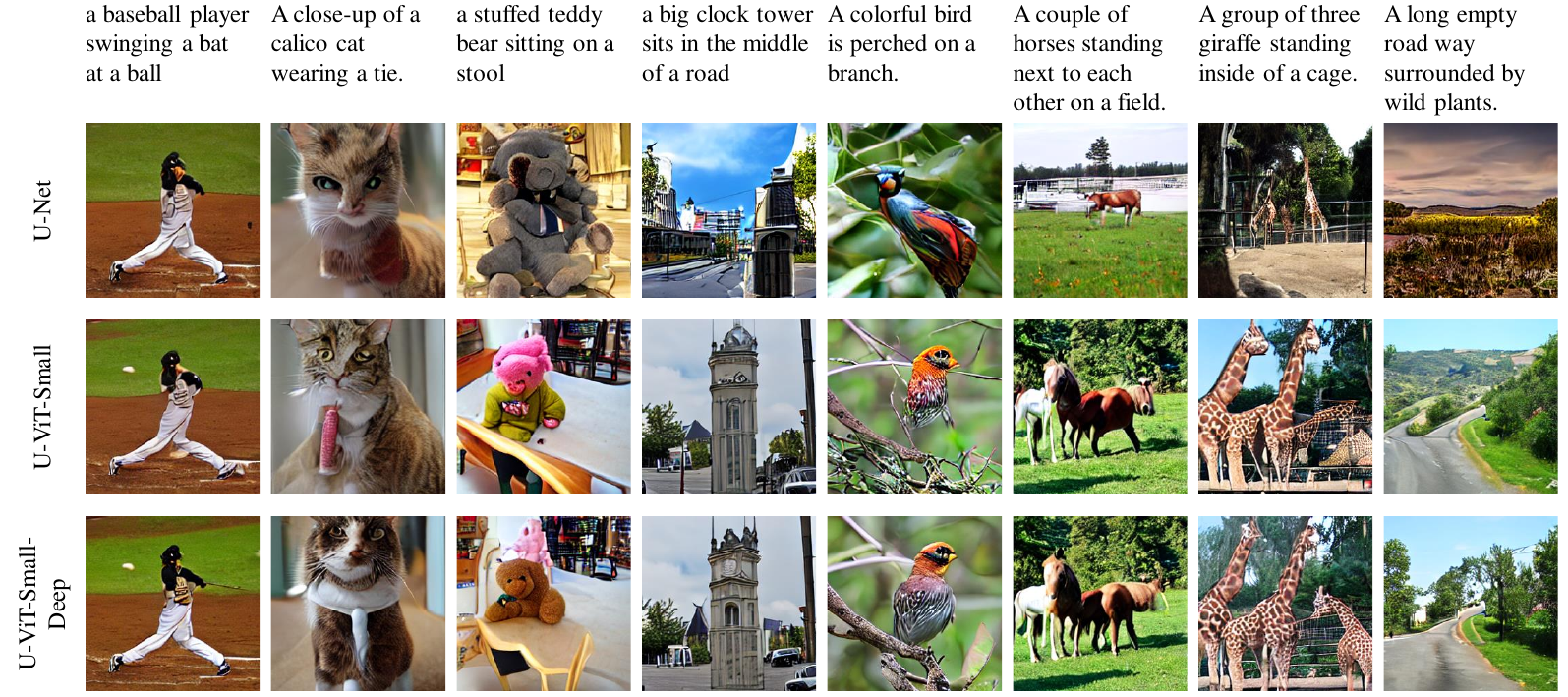}
    \caption{Text-to-image generation on MS-COCO. All the other settings except the backbone are the same. U-Net and U-ViT generate samples using the same random seed for a fair comparison. The random seed is unselected.}
    \label{fig:t2i}
\end{figure*}

\subsection{Unconditional and Class-Conditional Image Generation}
\label{sec:gen}

We compare U-ViT with prior diffusion models based on U-Net. We also compare with GenViT~\cite{yang2022your}, a smaller ViT which does not employ long skip connections, and incorporates time before normalization layers. Consistent with previous literature, we report the FID score~\cite{heusel2017gans} on 50K generated samples to measure the image quality.

As shown in Table~\ref{tab:fid}, U-ViT is comparable to U-Net on unconditional CIFAR10 and CelebA 64$\times$64, and meanwhile performs much better than GenViT. 

On class-conditional ImageNet 64$\times$64, we initially try the U-ViT-M configuration with 131M parameters. As shown in Table~\ref{tab:fid}, it gets a FID of 5.85, which is better than 6.92 of IDDPM that employs a U-Net with 100M parameters. To further improve the performance, we employ the U-ViT-L configuration with 287M parameters, and the FID improves from 5.85 to 4.26.

Meanwhile, we find that our U-ViT performs especially well in the latent space~\cite{rombach2022high}, where images are firstly converted to their latent representations before applying diffusion models. On class-conditional ImageNet 256$\times$256, our U-ViT obtains a state-of-the-art FID of 2.29, which outperforms all prior diffusion models. Table~\ref{tab:steps} further demonstrates that our U-ViT outperforms LDM under different number of sampling steps using the same sampler.
Note that our U-ViT also outperforms VQ-Diffusion, which is a discrete diffusion model~\cite{austin2021structured} that employs a transformer as its backbone. We also try replace our U-ViT with a U-Net with a similar amount of parameters and computational cost, where our U-ViT still outperforms U-Net (see details in Appendix~\ref{app:cmp}).
On class-conditional ImageNet 512$\times$512, our U-ViT outperforms ADM-G that directly models the pixels of images. In Figure~\ref{fig:samples}, we provide selected samples on ImageNet 256$\times$256 and ImageNet 512$\times$512, and random samples on other datasets, which have good quality and clear semantics. We provide more generated samples including class-conditional and random ones in Appendix~\ref{app:more}.

In Section~\ref{sec:imp} we have demonstrated the importance of long skip connection on small-scale dataset (i.e., CIFAR10). Figure~\ref{fig:skip_im} further shows it is also critical for large-scale dataset such as ImageNet.

In Appendix~\ref{app:metrics}, we present results of other metrics (e.g., sFID, inception score, precision and recall) as well as the computational cost (GFLOPs) with more U-ViT configurations on ImageNet. Our U-ViT is still comparable to state-of-the-art diffusion models on other metrics, and meanwhile has comparable if not smaller GFLOPs.

\subsection{Text-to-Image Generation on MS-COCO}
\label{sec:t2i}

We evaluate U-ViT for text-to-image generation on the standard benchmark dataset MS-COCO. We train our U-ViT in the latent space of images~\cite{rombach2022high} as detailed in Section~\ref{sec:setup}. We also train another latent diffusion model that employs a U-Net of comparable model size to U-ViT-S, and leave other parts unchanged. Its hyperparameters and training details are provided in Appendix~\ref{app:unet}. We report the FID score~\cite{heusel2017gans} to measure the image quality. Consistent with previous literature, we randomly draw 30K prompts from the MS-COCO validation set, and generate samples on these prompts to compute FID.

As shown in Table~\ref{tab:fid_ms}, our U-ViT-S already achieves a state-of-the-art FID among methods without accessing large external datasets during the training of generative models. By further increasing the number of layers from 13 to 17, our U-ViT-S (Deep) can even achieve a better FID of 5.48. Figure~\ref{fig:t2i} shows generated samples of U-Net and U-ViT using the same random seed for a fair comparison. We find U-ViT generates more high quality samples, and meanwhile the semantics matches the text better. For example, given the text \texttt{``a baseball player swinging a bat at a ball''}, U-Net generates neither the bat nor the ball. In contrast, our U-ViT-S generates the ball with even a smaller number of parameters, and our U-ViT-S (Deep) further generates the bat. We hypothesize this is because texts and images interact at every layer in our U-ViT, which is more frequent than U-Net that only interact at cross attention layer. We provide more samples in Appendix~\ref{app:more}.

\section{Conclusion}

This work presents U-ViT, a simple and general ViT-based architecture for image generation with diffusion models. U-ViT treats all inputs including the time, condition and noisy image patches as tokens and employs long skip connections between shallow and deep layers. We evaluate U-ViT in tasks including unconditional and class-conditional image generation, as well as text-to-image generation. Experiments demonstrate U-ViT is comparable if not superior to a CNN-based U-Net of a similar size. 
These results suggest that, for diffusion-based image modeling, the long skip connection is crucial while the down/up-sampling operators in CNN-based U-Net are not always necessary. We believe that U-ViT can provide insights for future research on backbones in diffusion models and benefit generative modeling on large scale cross-modality datasets.

\section*{Acknowledgments}

This work was supported by NSF of China Projects (Nos. 62061136001, 61620106010, 62076145, U19B2034, U1811461, U19A2081, 6197222);  Beijing Outstanding Young Scientist Program NO. BJJWZYJH012019100020098; a grant from Tsinghua Institute for Guo Qiang; the High Performance Computing Center, Tsinghua University; the Fundamental Research Funds for the Central Universities, and the Research Funds of Renmin University of China (22XNKJ13). C. Li was also sponsored by Beijing Nova Program. J.Z was also supported by the XPlorer Prize.

%%%%%%%%% REFERENCES
{\small
\bibliographystyle{ieee_fullname}
\bibliography{egbib}
}

\appendix

\onecolumn

\section{Experimental Setup}
\label{app:setup}

We list the experimental setup for U-ViT presented in the main paper in Table~\ref{tab:setting}.

\begin{table}[H]
\begin{center}
\scalebox{0.88}{
\begin{tabular}{lcccccc}
\toprule
    Dataset & CIFAR10 & CelebA 64$\times$64 & ImageNet 64$\times$64 & ImageNet 256$\times$256 & ImageNet 512$\times$512 & MS-COCO \\
    \midrule
    Latent space & $\times$ & $\times$ & $\times$ & $\checkmark$ & $\checkmark$ & $\checkmark$ \\
    Latent shape & - & - & - & 32$\times$32$\times$4 & 64$\times$64$\times$4 & 32$\times$32$\times$4 \\
    Image decoder & - & - & - & ft-EMA & ft-EMA & original \\
    \midrule
    Batch size & 128 & 128 & 1024 & 1024 & 1024 & 256 \\
    Training iterations & 500K & 500K & 300K & 500K & 500K & 1M \\
    Warm-up steps & 2.5K & 5K & 5K & 5K & 5K & 5K \\
    \midrule
    Optimizer & AdamW & AdamW & AdamW & AdamW & AdamW & AdamW \\
    Learning rate & 2e-4 & 2e-4 & 3e-4 & 2e-4 & 2e-4 & 2e-4 \\
    Weight decay & 0.03 & 0.03 & 0.03 & 0.03 & 0.03 & 0.03 \\
    Betas & (0.99, 0.999) & (0.99, 0.99) & (0.99, 0.99) & (0.99, 0.99) & (0.99, 0.99) & (0.9, 0.9) \\
    \midrule
    Noise schedule & VP & VP & VP & SD & SD & SD \\
    \midrule
    Sampler & EM & EM & DPM-Solver & DPM-Solver & DPM-Solver & DPM-Solver \\
    Sampling steps & 1K & 1K & 50 & 50 & 50 & 50 \\
    \midrule
    CFG & $\times$ & $\times$ & $\times$ & $\checkmark$ & $\checkmark$ & $\checkmark$ \\
    $p_{\mathrm{uncond}}$ & - & - & - & 0.1 & 0.1 & 0.1 \\
    Guidance strength & - & - & - & 0.4 & 0.7 & 1 \\
    \midrule
    Convolution & $\checkmark$ & $\checkmark$ & $\checkmark$ & $\times$ & $\times$ & $\checkmark$\\
    \bottomrule
\end{tabular}}\vspace{-.2cm}
\end{center}
\caption{The experimental setup for U-ViT in the main paper. ``ft-EMA" and ``original" correspond to different weights of the image decoder provided in \url{https://huggingface.co/stabilityai/sd-vae-ft-ema}. ``VP" represents the continuous-time variance preserving noise schedule~\cite{song2020score}. ``SD" represents the discrete-time noise schedule used in Stable Diffusion~\cite{rombach2022high}. ``EM" represents the Euler-Maruyama SDE sampler~\cite{song2020score}. ``DPM-Solver" represents the DPM-Solver ODE sampler~\cite{lu2022dpm}. ``$p_{\mathrm{uncond}}$" represents the unconditional training probability in classifier free guidance (CFG). ``Convolution" represents whether to add a 3$\times$3 convolutional block before output.}
\label{tab:setting}
\end{table}

In our early experiments, we try learning rates between 1e-4 and 5e-4, and find that a learning rate of 2e-4 performs well for all datasets. On ImageNet 64$\times$64, a learning rate of 3e-4 could further improve the performance. We try weight decay between 0.01 and 0.05, and find that a weight decay of 0.03 performs well for all datasets. We try the running coefficients $\beta_1$, $\beta_2$ of AdamW among $\{0.9, 0.99, 0.999\}$, and find that $(\beta_1, \beta_2) = (0.99, 0.99)$ performs well for all datasets. On CIFAR10, $\beta_2 = 0.999$ could further improve the performance. On MS-COCO, $(\beta_1, \beta_2) = (0.9, 0.9)$ could further improve the performance.
We train with mixed precision for efficiency, and the training time and devices are listed in Table~\ref{tab:time}. Besides, the training memory of U-ViT can be greatly reduced with the gradient checkpointing trick. For example, the memory for forward and backward on a single A100 can be reduced from 53GB to 10GB when training U-ViT-L/2 with a batch size of 128 on ImageNet 256$\times$256.

During inference, with 1 A100, generating 500 samples with DPM-Solver takes around 19 seconds, 34 seconds, 59 seconds, 89 seconds, with U-ViT-S, U-ViT-M, U-ViT-L, U-ViT-H respectively. The time would double if classifier-free guidance is used.

\begin{table}[H]
    \centering
    \scalebox{0.95}{
    \begin{tabular}{ccccc}
    \toprule
    Dataset & Model & Training devices & Training time & Training iterations \\
    \midrule
    CIFAR10 & U-ViT-S/2 & 4 GeForce RTX 2080 Ti & 24 hours & 500K \\
    \midrule
    CelebA & U-ViT-S/4 & 4 GeForce RTX 2080 Ti & 24 hours & 500K \\
    \midrule
    ImageNet 64$\times$64 & U-ViT-M/4 & 8 A100 & 59 hours & 300K \\
    ImageNet 64$\times$64 & U-ViT-L/4 & 8 A100 & 100 hours & 300K \\
    \midrule
    ImageNet 256$\times$256 & U-ViT-L/2 & 8 A100 & 100 hours & 300K \\
    ImageNet 256$\times$256 & U-ViT-H/2 & 8 A100 & 208 hours & 500K \\
    \midrule
    ImageNet 512$\times$512 & U-ViT-L/4 & 8 A100 & 166 hours & 500K \\
    ImageNet 512$\times$512 & U-ViT-H/4 & 8 A100 & 208 hours & 500K \\
    \midrule
    MS-COCO & U-ViT-S/2 & 4 A100 & 60 hours & 1M \\
    MS-COCO & U-ViT-S/2 (deep) & 4 A100 & 74 hours & 1M \\
    \bottomrule
    \end{tabular}}
    \caption{The training devices and time.}
    \label{tab:time}
\end{table}

\section{Details of the U-Net Baseline on MS-COCO}
\label{app:unet}

We employ the U-Net with cross attention provided by LDM~\cite{rombach2022high} for the baseline. The U-Net is performed on the 32$\times$32 resolution latent representation, and down-samples it to 16$\times$16, 8$\times$8 and 4$\times$4 resolution. The number of channels is 128 at 32$\times$32 resolution, and 256 at other resolutions. Each resolution has 2 residual blocks. The U-Net performs self attention and cross attention at 16$\times$16 and 8$\times$8 resolution. Such a configuration leads to a total of 53M parameters, which is comparable to 45M of U-ViT-Small for a fair comparison. We use the AdamW optimizer with weight decay set to 0.01 and running coefficients $\beta_1$, $\beta_2$ set to (0.9, 0.999), which are the setting used across LDM~\cite{rombach2022high}. We tune the learning rate of U-Net and find 2e-4 performs the best. The training iterations and the batch size of U-Net are the same to U-ViT for a fair comparison.

\section{Results of Other Metrics and Configurations on ImageNet}
\label{app:metrics}
We present results of FID~\cite{heusel2017gans}, sFID~\cite{nash2021generating}, inception score (IS)~\cite{salimans2016improved}, precision and recall~\cite{kynkaanniemi2019improved} on ImageNet in Table~\ref{tab:metrics}. Our U-ViT is still comparable to state-of-the-art diffusion models based on U-Net on these metrics, and meanwhile has comparable if not smaller GFLOPs.

\begin{table}[H]
\centering
\scalebox{0.66}{\begin{tabular}{lccrrrrr}
\toprule
    \textbf{ImageNet 64$\times$64} & \#Params & GFLOPs & FID$\downarrow$ & sFID$\downarrow$ & IS$\uparrow$ & Precision$\uparrow$ & Recall$\uparrow$ \\
\midrule
    ADM~\cite{dhariwal2021diffusion} & 296M & 110 & 2.07 & 4.29 & - & 0.74 & 0.63  \\
\arrayrulecolor{black!30}\midrule
    U-ViT-M/4 \small{(VP, trained 300K, w/ conv)} & 131M & 35 & 5.85 & 4.09 & 33.71 & 0.69 & 0.61 \\
    U-ViT-L/4 \small{(VP, trained 300K, w/ conv)} & 287M & 77 & 4.26 & 3.77 & 40.66 & 0.71 & 0.62 \\
\arrayrulecolor{black}\bottomrule
\\
\toprule
    \textbf{ImageNet 256$\times$256} & \#Params & GFLOPs & FID$\downarrow$ & sFID$\downarrow$ & IS$\uparrow$ & Precision$\uparrow$ & Recall$\uparrow$ \\
\midrule
    ADM-G, ADM-U~\cite{dhariwal2021diffusion} & 296M + 65M (Cls) + 312M (SR) & 110 + 19 (Cls) + 632 (SR)  & 3.94 & 6.14 & 215.84 & 0.83 & 0.53 \\
    LDM~\cite{rombach2022high} & 400M + 55M (AE) & 104 + 336 (AE) & 3.60 & - & 247.67 & 0.87 & 0.48 \\
\arrayrulecolor{black!30}\midrule
    U-ViT-L/2 \small{(VP, trained 300K, w/ conv, original, $p_{\mathrm{uncond}}$=0.15)} & 287M + 84M (AE) & 77 + 312 (AE) & 3.40 & 6.63 & 219.94 & 0.83 & 0.52 \\
    U-ViT-H/2 \small{(VP, trained 300K, w/ conv, original, $p_{\mathrm{uncond}}$=0.1)} & 501M + 84M (AE) & 133 + 312 (AE) & 3.10 & 6.70 & 250.82 & 0.84 & 0.53\\
    U-ViT-H/2 \small{(VP, trained 300K, w/o conv, original, $p_{\mathrm{uncond}}$=0.1)} & 501M + 84M (AE) & 133 + 312 (AE) & 3.74 &  8.04 & 244.47 & 0.84 & 0.51\\
\arrayrulecolor{black!30}\midrule
    U-ViT-H/2 \small{(SD, trained 300K, w/ conv, original, $p_{\mathrm{uncond}}$=0.1)} & 501M + 84M (AE) & 133 + 312 (AE) & 3.14 & 7.81 & 229.03 & 0.82 & 0.55\\
    U-ViT-H/2 \small{(SD, trained 300K, w/o conv, original, $p_{\mathrm{uncond}}$=0.15)} & 501M + 84M (AE) & 133 + 312 (AE) & 2.90 & 7.70 & 242.59 & 0.81 & 0.56 \\
    U-ViT-H/2 \small{(SD, trained 300K, w/o conv, original, $p_{\mathrm{uncond}}$=0.1)} & 501M + 84M (AE) & 133 + 312 (AE) & 2.78 & 7.55 & 251.83 & 0.82 & 0.56\\
    U-ViT-H/2 \small{(SD, trained 500K, w/o conv, original, $p_{\mathrm{uncond}}$=0.1)} & 501M + 84M (AE) & 133 + 312 (AE) & 2.65 & 8.17 & 260.34 & 0.81 & 0.57\\
    U-ViT-H/2 \small{(SD, trained 500K, w/o conv, ft-EMA, $p_{\mathrm{uncond}}$=0.1)} & 501M + 84M (AE) & 133 + 312 (AE) & 2.29 & 5.68 & 263.88 & 0.82 & 0.57\\
\arrayrulecolor{black}\bottomrule
\\
\toprule
    \textbf{ImageNet 512$\times$512} & \#Params & GFLOPs & FID$\downarrow$ & sFID$\downarrow$ & IS$\uparrow$ & Precision$\uparrow$ & Recall$\uparrow$ \\
\midrule
    ADM-G, ADM-U~\cite{dhariwal2021diffusion} & 422M + 43M (Cls) + 309M (SR) & 307 + 21 (Cls) + 2506 (SR) & 3.85 & 5.86 & 221.72 & 0.84 & 0.53\\
\arrayrulecolor{black!30}\midrule
    U-ViT-L/4 \small{(VP, trained 500K, w/ conv, original, $p_{\mathrm{uncond}}$=0.15)} & 287M + 84M (AE) & 77 + 1260 (AE) & 4.67 & 5.87 & 213.28 & 0.87 &  0.45 \\
\arrayrulecolor{black!30}\midrule
    U-ViT-H/4 \small{(SD, trained 500K, w/o conv, original, $p_{\mathrm{uncond}}$=0.1)} & 501M + 84M (AE) & 133 + 1260 (AE) & 4.34 & 8.44 & 261.13 & 0.84 & 0.48\\
    U-ViT-H/4 \small{(SD, trained 500K, w/o conv, ft-EMA, $p_{\mathrm{uncond}}$=0.1)} & 501M + 84M (AE) & 133 + 1260 (AE) & 4.05 & 6.44 & 263.79 & 0.84 & 0.48\\
\arrayrulecolor{black}\bottomrule
\end{tabular}}
    \caption{Results of FID~\cite{heusel2017gans}, sFID~\cite{nash2021generating}, inception score (IS)~\cite{salimans2016improved}, precision and recall~\cite{kynkaanniemi2019improved} on ImageNet. We also show the number of parameters as well as the GFLOPs.}
    \label{tab:metrics}
\end{table}

\section{CKA Analysis}
\label{app:cka}
% within and across models 

Centered kernel alignment (CKA) is widely used to analyze similarity between hidden representations in deep neural networks~\cite{raghu2021vision,kornblith2019similarity,cortes2012algorithms}. In this section, we use the CKA method to analyze hidden representations of networks that employ three ways to combine long skip branches: (1) concatenation, i.e., $\mathrm{Linear}(\mathrm{Concat}(\vh_m, \vh_s))$; (2) addition, i.e., $\vh_m + \vh_s$; (3) no long skip connection. These three ways are elaborated in Section 3.1 in the main paper. 
We evaluate hidden representations after each transformer block and fix the input time as $t=0.5$ on CIFAR10.

\begin{figure}[H]
\centering
\subfloat[Concatenation]{\includegraphics[width=0.32\linewidth]{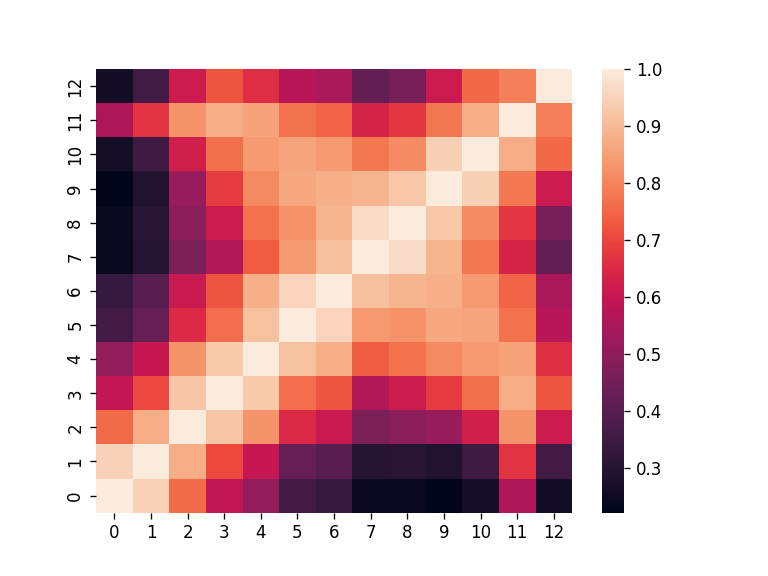}}
\hspace{.1cm}
\subfloat[Addition]{\includegraphics[width=0.32\linewidth]{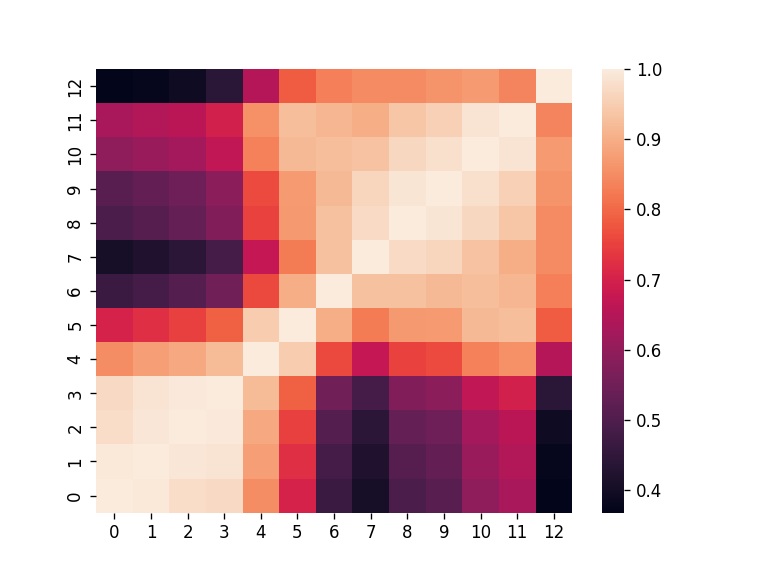}}
\hspace{.1cm}
\subfloat[No long skip connection]{\includegraphics[width=0.32\linewidth]{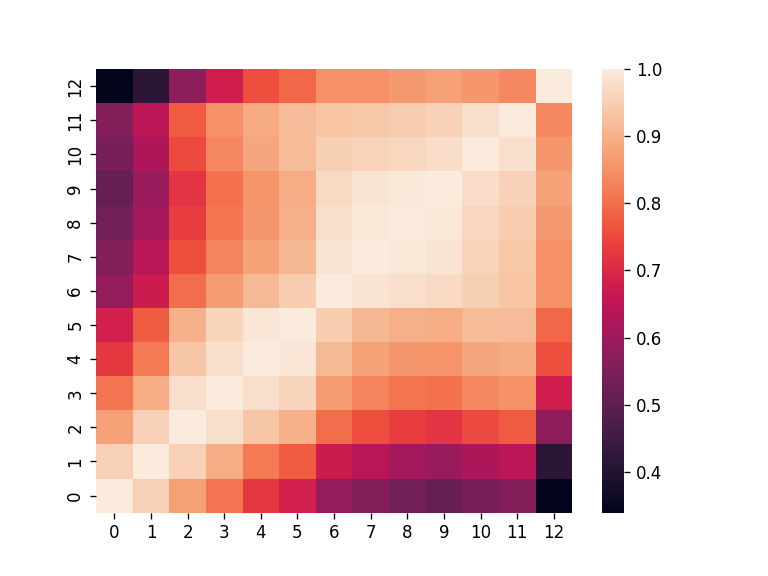}}
\caption{CKA analysis on hidden representations of networks that employ three ways to combine long skip branches. We analyze the similarity between hidden representations after each transformer block in the same network.}
\label{fig:cka-v1}
\end{figure}

We find that the ``addition'' and ``no long skip connection'' settings share a similar phenomenon that neighboring blocks in the network have similar representations, e.g., blocks 0-3, 6-11 in Figure~{\ref{fig:cka-v1} (b)}, and blocks 0-5, 6-11 in Figure~{\ref{fig:cka-v1} (c)}. In contrast, the representations of neighboring blocks under the ``concatenation'' setting have low similarity, as shown in Figure~{\ref{fig:cka-v1} (a)}. Thus, the ``concatenation'' setting significantly changes the representations in the transformer, while the ``addition'' setting does not.

\section{Compare with U-Net Under Similar Amount of Parameters and Computational Cost}
\label{app:cmp}

\begin{figure}[H]
    \centering
    \includegraphics[width=0.4\linewidth]{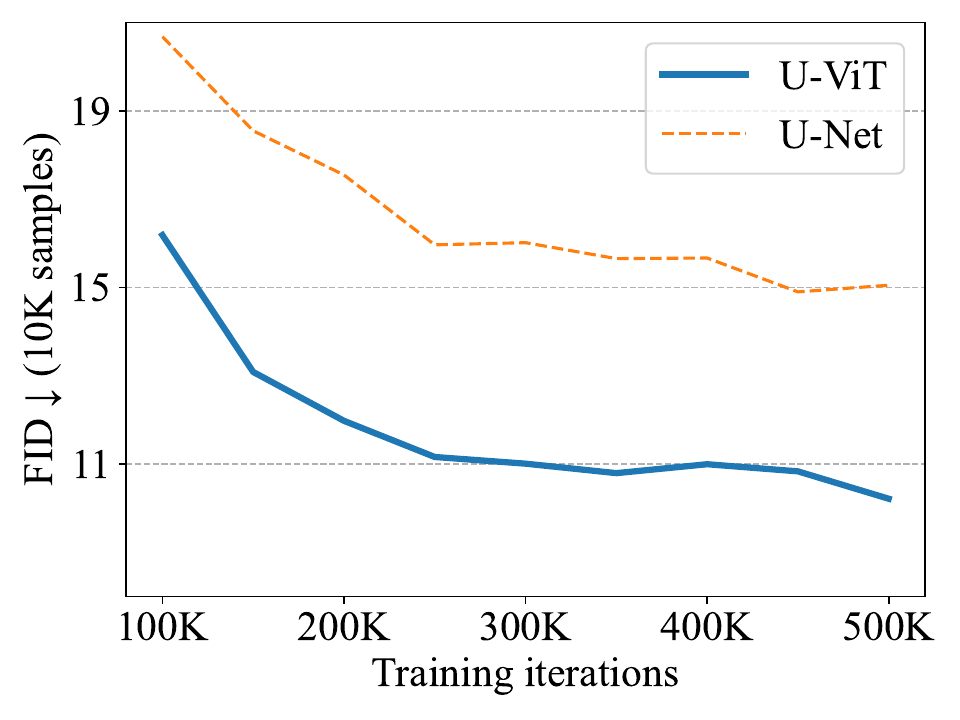}
    \caption{Compare with U-Net under similar amount of parameters and computational cost (w/o classifier-free guidance).}
    \label{fig:cmp_uvit_unet}
\end{figure}

On ImageNet 256$\times$256, we also try replace our U-ViT with a U-Net with a similar amount of parameters and computational cost. The U-Net employs implementation from ADM~\cite{dhariwal2021diffusion}. We set the model channels as 320, the channel multiplier as (2, 2, 4), the number of residual blocks as 3, and employs attention at 2$\times$ and 4$\times$ down-sampling. This leads to a U-Net of 646M parameters and 135 GFLOPs, and our U-ViT has 501M parameters and 133 GFLOPs. We use the same optimizer configuration as ADM. As shown in Figure~\ref{fig:cmp_uvit_unet}, our U-ViT consistently outperforms U-Net at different training iterations without classifier-free guidance. We also evaluate FID with 50K samples at 500K training iterations. With no classifier-free guidance, U-ViT obtains a FID of 6.58 and U-Net obtains a FID of 10.69. With a classifier-free guidance scale of 0.4, U-ViT obtains a FID of 2.29 and U-Net obtains a FID of 2.66. Under both settings, our U-ViT outperforms U-Net.

\clearpage
\section{Additional Samples}
\label{app:more}

\begin{figure}[ht]
    \centering
    \includegraphics[width=0.7\linewidth]{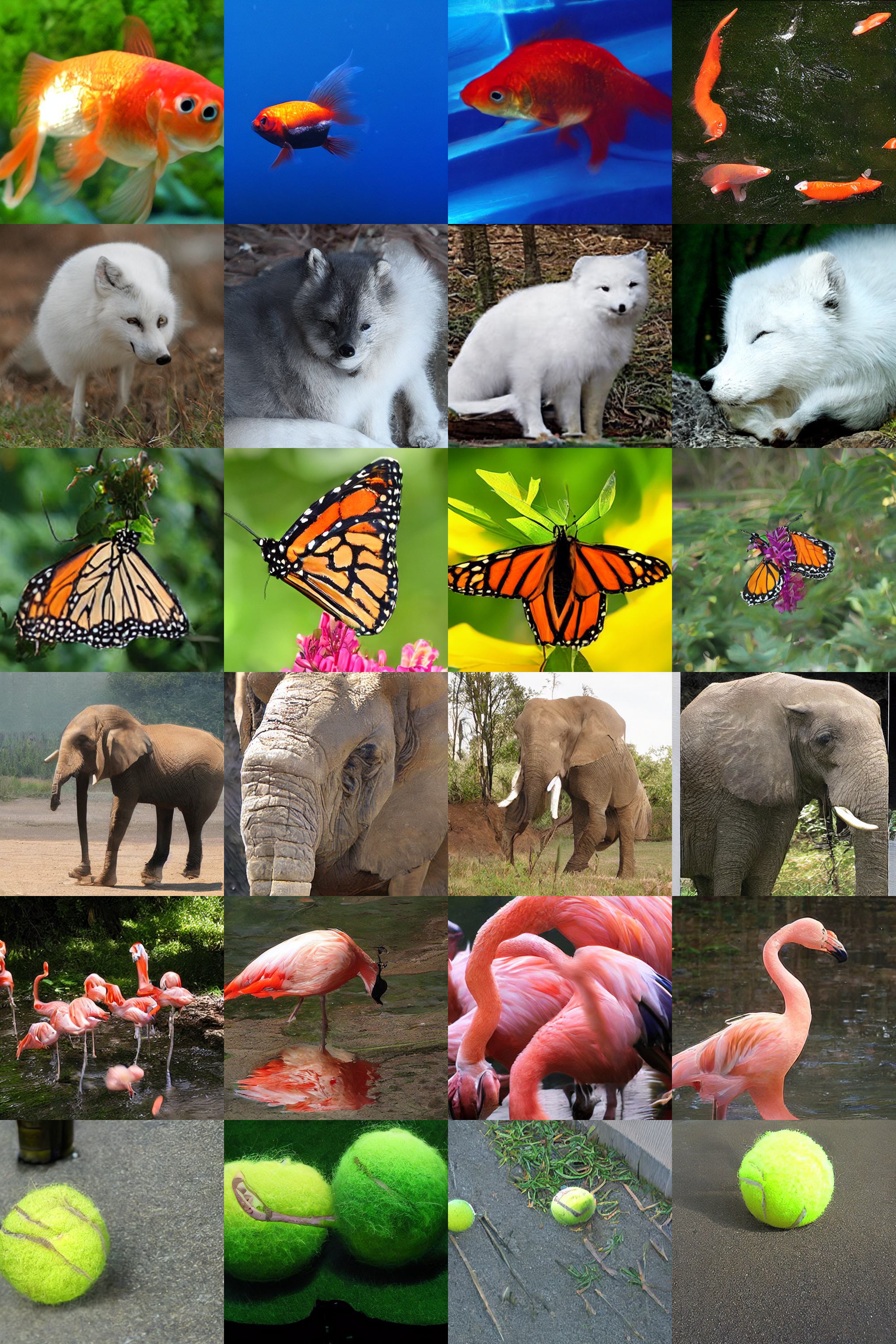}
    \caption{Generated samples on ImageNet 512$\times$512, conditioned on goldfish (1), arctic fox (279), monarch butterfly (323), african elephant (386), flamingo (130), tennis ball (852).}
    \label{fig:cc_imagenet512_1}
\end{figure}

\begin{figure}[ht]
    \centering
    \includegraphics[width=0.7\linewidth]{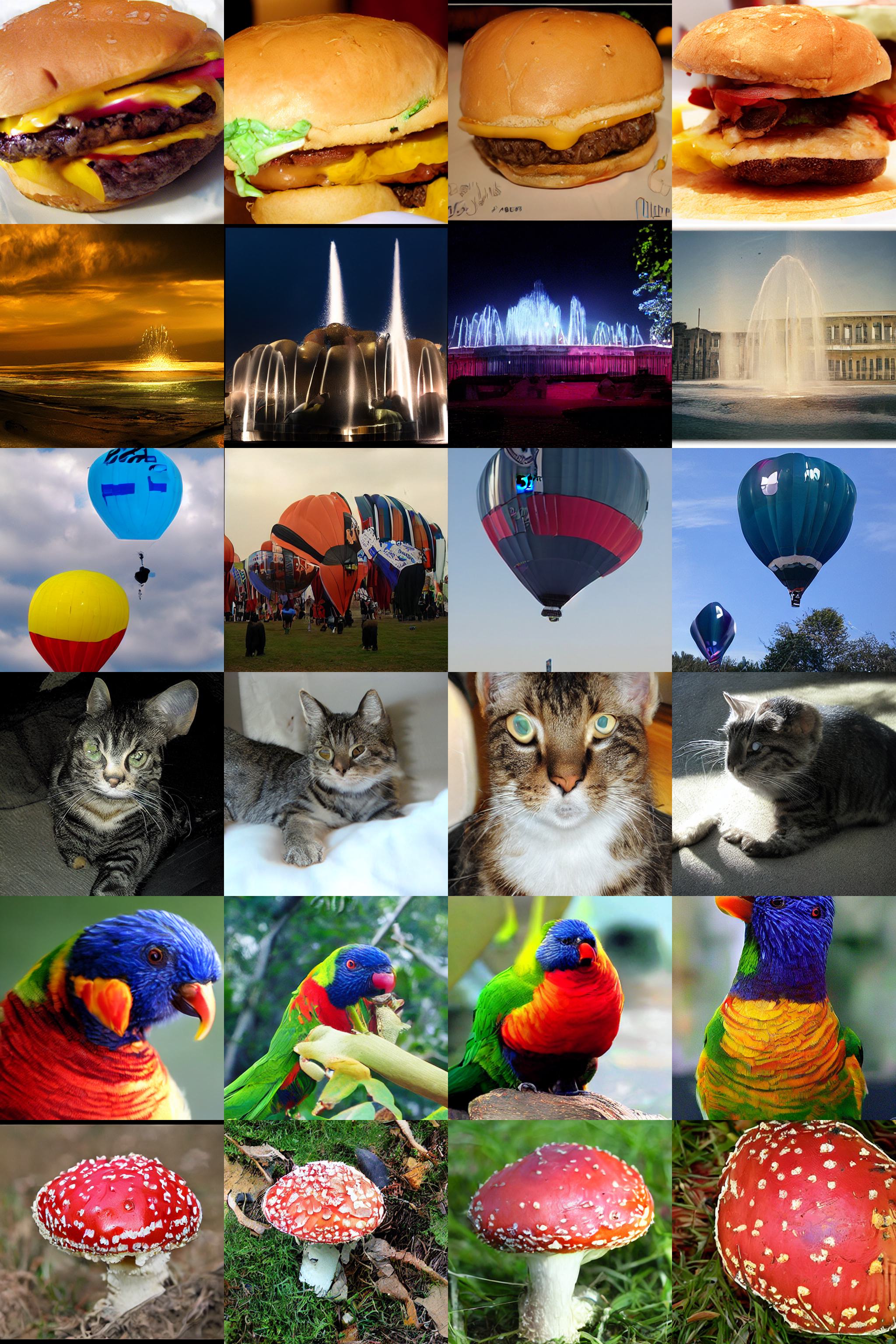}
    \caption{Generated samples on ImageNet 512$\times$512, conditioned on cheeseburger (933), fountain (562), balloon (417), tabby cat (281), lorikeet (90), agaric (992).}
    \label{fig:cc_imagenet512_2}
\end{figure}

\begin{figure}[ht]
    \centering
    \includegraphics[width=0.7\linewidth]{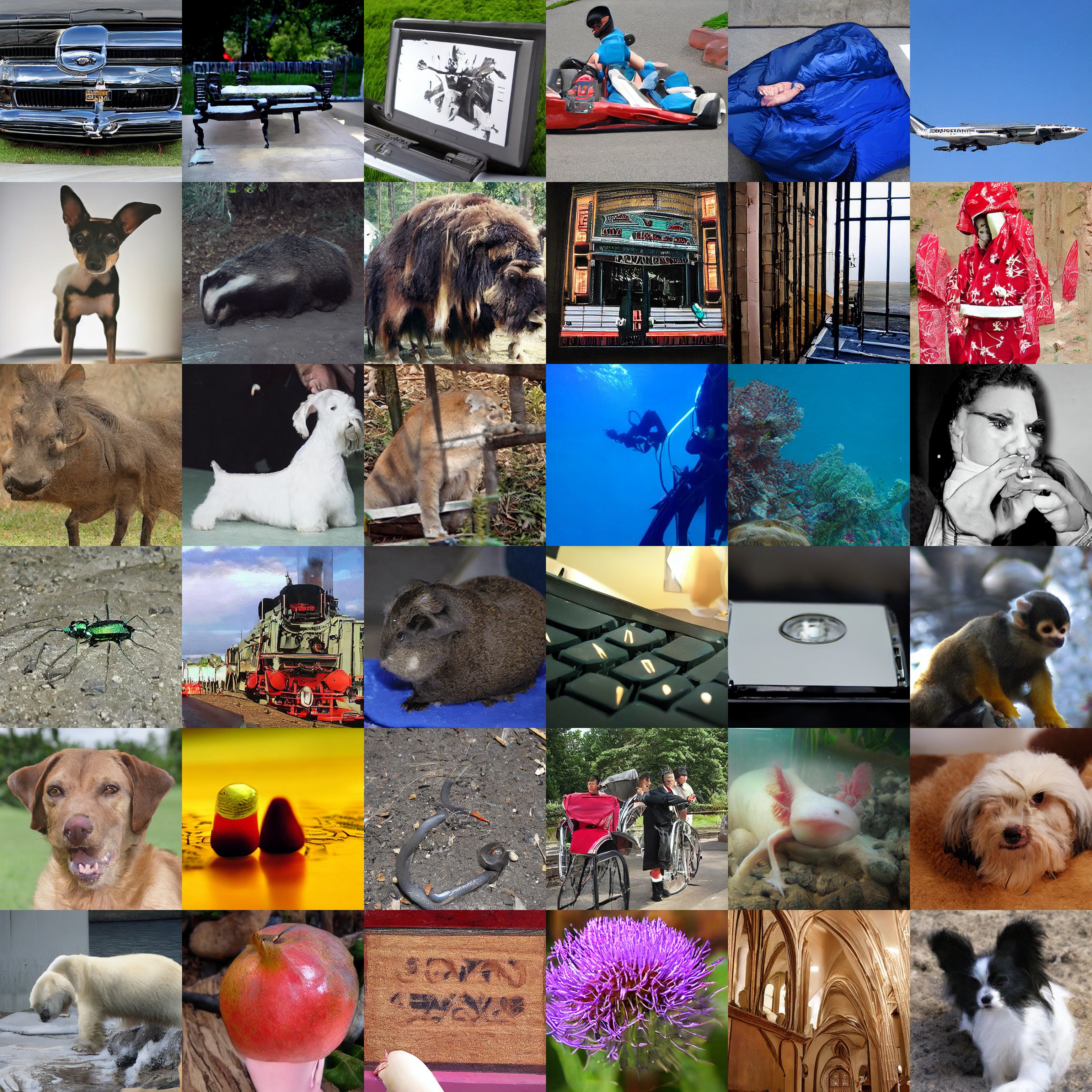}
    \caption{Random samples on ImageNet 512$\times$512.}
    \label{fig:random_imagenet512}
\end{figure}

\begin{figure}[ht]
    \centering
    \includegraphics[width=0.7\linewidth]{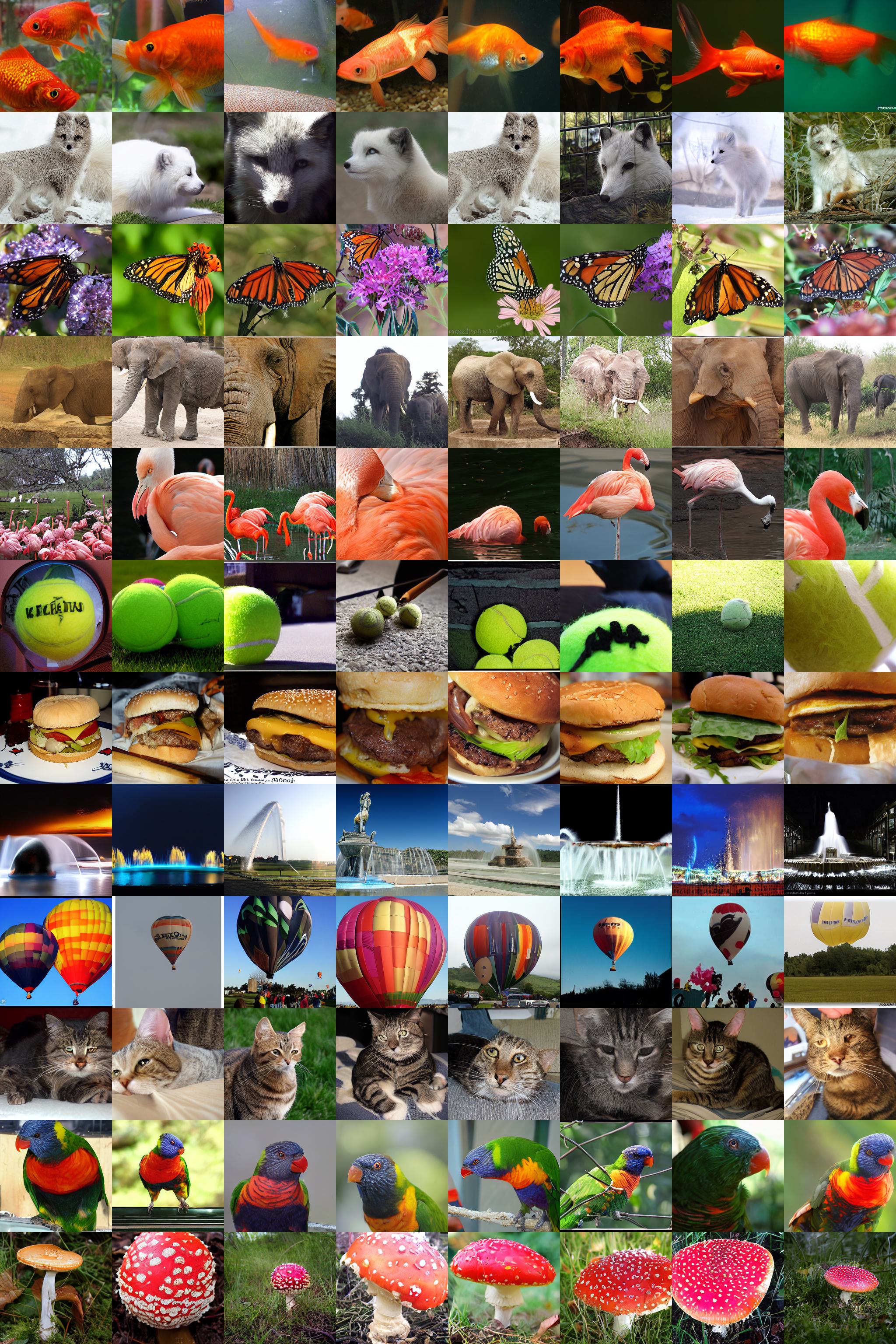}
    \caption{Generated samples on ImageNet 256$\times$256, conditioned on goldfish (1), arctic fox (279), monarch butterfly (323), african elephant (386), flamingo (130), tennis ball (852), cheeseburger (933), fountain (562), balloon (417), tabby cat (281), lorikeet (90), agaric (992).}
    \label{fig:cc_imagenet256}
\end{figure}

\begin{figure}[ht]
    \centering
    \includegraphics[width=0.7\linewidth]{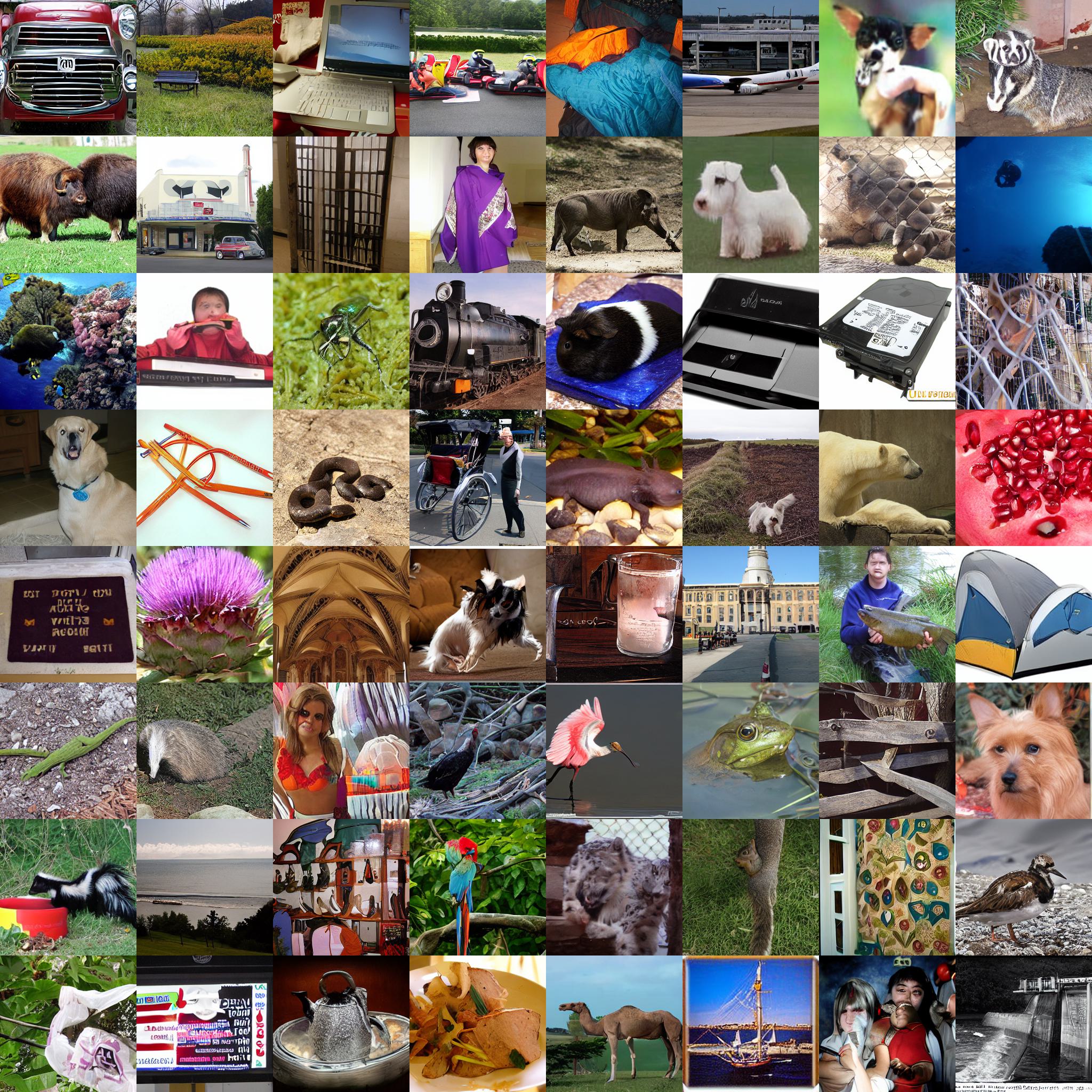}
    \caption{Random samples on ImageNet 256$\times$256.}
    \label{fig:random_imagenet256}
\end{figure}

\begin{figure}[ht]
    \centering
    \includegraphics[width=\linewidth]{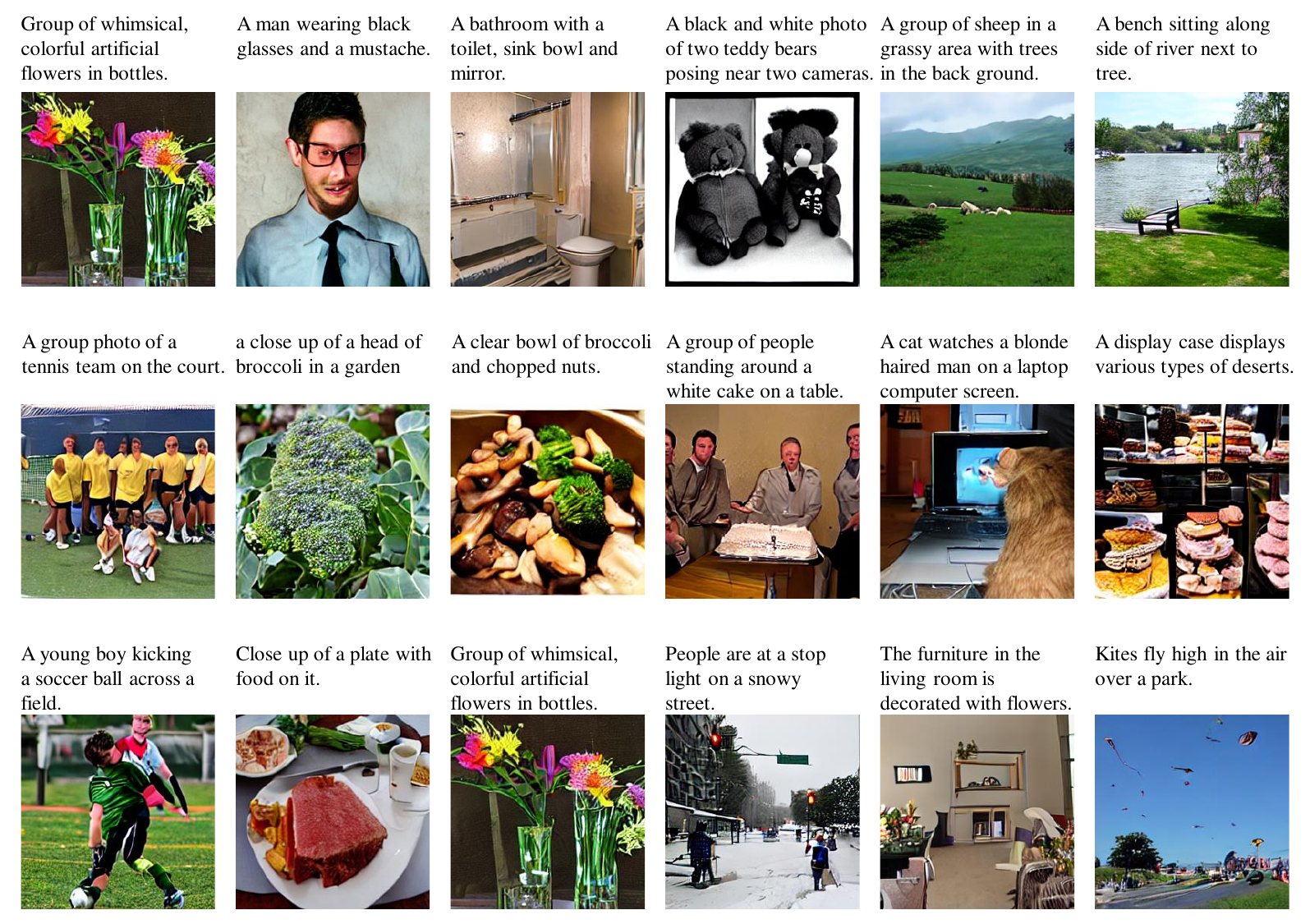}
    \caption{Random samples on MS-COCO. Prompts are randomly drawn from the validation set.}
    \label{fig:mscoco}
\end{figure}

\end{document}